\documentclass{article}
\usepackage[preprint,nonatbib]{neurips_2021}
\usepackage{amsmath}
\usepackage{mathtools}
\usepackage[bb=boondox,bbscaled=.95]{mathalfa}
\usepackage{tikz}
\usepackage{pgfplots}
\usepackage{physics}
\usepackage{bm}
\usepackage{thmtools, thm-restate}

\usepgfplotslibrary{groupplots}
\usepgfplotslibrary{patchplots}
\pgfplotsset{compat=1.14}

\DeclareMathOperator{\dom}{dom}
\DeclareMathOperator{\vect}{vec}
\DeclareMathOperator{\R}{\mathbb{R}}
\DeclareMathOperator{\Nat}{\mathbb{N}}

\DeclareMathOperator{\cG}{\mathcal{G}}
\DeclareMathOperator{\cK}{\mathcal{K}}
\DeclareMathOperator{\cS}{\mathcal{S}}
\DeclareMathOperator{\cT}{\mathcal{T}}

\newcommand{\nT}{{n_\theta}}
\newcommand{\nZ}{{n_z}}
\newcommand{\nX}{{n_x}}
\newcommand{\nY}{{n_y}}

\DeclareMathOperator{\V}{\mathcal{V}}
\DeclareMathOperator{\D}{\mathcal{D}}

\DeclareMathOperator{\N}{\mathcal{N}}
\DeclareMathOperator{\K}{\mathcal{K}}
\DeclareMathOperator{\G}{\mathcal{G}}
\DeclareMathOperator{\E}{\mathcal{E}}
\DeclareMathOperator{\I}{\mathcal{I}}

\newcommand{\x}{\mathbf{x}}

\newcommand{\yb}{\mathbf{y}}
\newcommand{\y}{\mathbf{y}}

\newcommand{\lc}{\left(}
\newcommand{\rc}{\right)}

\tikzset{
  NNnode/.pic={
  \pgfmathsetmacro\RecH{2}
  \pgfmathsetmacro\RecW{\RecH/10}
  \coordinate (-ll) at (-\RecW/2,-\RecH/2);
  \coordinate (-ur) at (\RecW/2,\RecH/2);
  \coordinate (-lr) at (-ll-|-ur);
  \coordinate (-ul) at (-ll|--ur);
  \path (-ul) -- (-ur) coordinate[midway] (-north);
  \path (-ll) -- (-lr) coordinate[midway] (-south);
  \path (-ll) -- (-ul) coordinate[midway] (-west);
  \path (-ur) -- (-lr) coordinate[midway] (-east);

  \begin{scope}[shift={(-\RecW/2,-\RecH/2)}]
  \draw (-ll) rectangle (-ur);
  \foreach \y in {0.05,0.5,0.75,0.85,0.95}
    \draw (0.5*\RecW,\RecH*\y) circle[radius=0.3*\RecW];
  \foreach \y in {0.275,0.625} {
    \fill (\RecW*0.4,\y*\RecH-0.1*\RecW) rectangle (0.6*\RecW,\y*\RecH-0.3*\RecW);
    \fill (\RecW*0.4,\y*\RecH+0.1*\RecW) rectangle (0.6*\RecW,\y*\RecH+0.3*\RecW);
    }
  \end{scope}
  }
}

\usepackage[most]{tcolorbox}

\definecolor{olive}{rgb}{0.6, 0.6, 0.2}
\definecolor{sand}{rgb}{0.8666666666666667, 0.8, 0.4666666666666667}
\definecolor{wine}{rgb}{0.5333333333333333, 0.13333333333333333, 0.3333333333333333}
\definecolor{deblue}{RGB}{11,132,147}
\definecolor{ocra}{RGB}{204, 119, 34}
\definecolor{depurple}{RGB}{131, 102, 135}
\definecolor{degrey}{RGB}{186, 172, 172}
\newcommand{\fcircle}[2][red,fill=red]{\tikz[baseline=-0.5ex]\draw[#1,radius=#2] (0,0.03) circle ;}
\newtcolorbox{CatchyBox}[2][]{
    lower separated=false,
    colback=white!80!degrey!90!depurple,
    colframe=white, fonttitle=\bfseries,
    colbacktitle=white!50!degrey!90!depurple,
    coltitle=black,
    enhanced,
    attach boxed title to top left={xshift=.02\linewidth,yshift=-4mm},
    title=#2,#1}
\usepackage{wrapfig}
\usepackage{enumitem}
\usepackage[colorlinks=true,linkcolor=blue,citecolor=blue]{hyperref}  
\usepackage[round]{natbib}
\usepackage{microtype}
\usepackage{graphicx}
\usepackage{booktabs, physics} 
\usepackage{amssymb} 
\usepackage[toc,page,header]{appendix}
\usepackage{minitoc}
\usepackage{hyperref}

\title{Continuous--Depth Neural Models for Dynamic Graph Prediction}
\author{Michael Poli$^{1,}$\thanks{Equal contribution. Author order was decided via coin flip. $^{1}$KAIST. $^{2}$The University of Tokyo $^{3}$Syntensor. Corresponding author: \textit{Michael Poli}, email: {\texttt{poli\_m@kaist.ac.kr}}}  , Stefano Massaroli$^{2,*}$, Clayton M. Rabideau$^3$\\ \textbf{Junyoung Park}$^1$, \textbf{Atsushi Yamashita}$^2$, \textbf{Hajime Asama}$^2$, \textbf{Jinkyoo Park}$^1$}

\begin{document}
\setlength{\abovedisplayskip}{1pt}
\setlength{\belowdisplayskip}{1pt}

\maketitle
\begin{abstract}
We introduce the framework of continuous--depth \textit{graph neural networks} (GNNs). \textit{Neural graph differential equations} (Neural GDEs) are formalized as the counterpart to GNNs where the input--output relationship is determined by a \textit{continuum} of GNN layers, blending discrete topological structures and differential equations. The proposed framework is shown to be compatible with static GNN models and is extended to dynamic and stochastic settings through hybrid dynamical system theory. Here, Neural GDEs improve performance by exploiting of the underlying dynamics geometry, further introducing the ability to accommodate irregularly sampled data. Results prove the effectiveness of the proposed models across applications, such as traffic forecasting or prediction in genetic regulatory networks.
\end{abstract}
\doparttoc
\faketableofcontents
\section{Introduction}
Introducing appropriate inductive biases on deep learning models is a well--known approach to improving sample efficiency and generalization performance \citep{battaglia2018relational}. Graph neural networks (GNNs) represent a general computational framework 
\begin{wrapfigure}[22]{r}{0.55\textwidth}
    \centering
    \includegraphics[width=\linewidth]{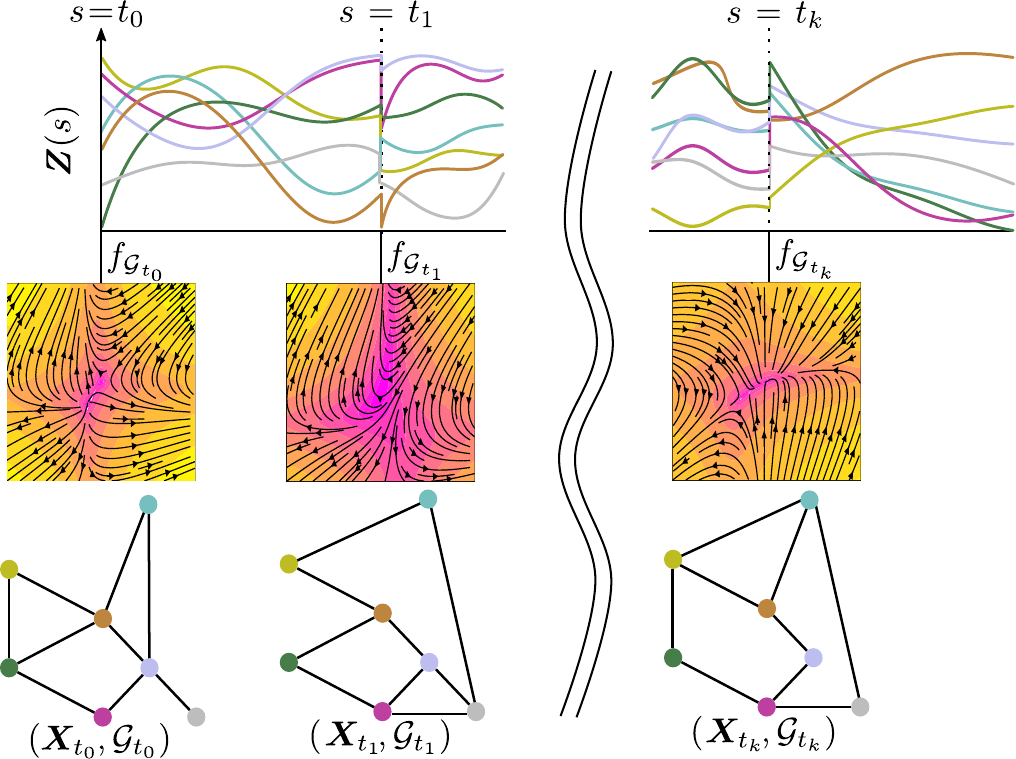}
    \caption{\textit{Neural graph differential equations} (Neural GDEs) model vector fields defined on graphs, both in cases when the structure is fixed or changes in time, via a continuum of \textit{graph neural network} (GNN) layers.}
    \label{fig:vek}
\end{wrapfigure}
for imposing such inductive biases when the problem structure can be encoded as a graph or in settings where prior knowledge about entities composing a target system can itself be described as a graph
\citep{li2018combinatorial,gasse2019exact,sanchez2018graph,you2019g2sat}. 
GNNs have shown remarkable results in various application areas such as node classification \citep{zhuang2018dual,gallicchio2019fast}, graph classification \citep{yan2018spatial} and forecasting \citep{li2017diffusion,wu2019graph} as well as generative tasks \citep{li2018learning,you2018graphrnn}.
A different but equally important class of inductive biases is concerned with the type of temporal behavior of the systems from which the data is collected i.e., discrete or continuous dynamics. Although deep learning has traditionally been a field dominated by discrete models, recent advances propose a treatment of neural networks equipped with a continuum of layers \citep{weinan2017proposal, chen2018neural, massaroli2020dissecting}. This view allows a reformulation of the forward and backward pass as the solution of the initial value problem of an \textit{ordinary differential equation} (ODE). The resulting \textit{continuous--depth} paradigm has successfully guided the discovery of novel deep learning models, with applications in prediction \citep{rubanova2019latent,greydanus2019hamiltonian}, control \citep{du2020model}, density estimation \citep{grathwohl2018ffjord,lou2020neural,mathieu2020riemannian}, time series classification \citep{kidger2020neural}, among others.
In this work we develop and experimentally validate a framework for the systematic blending of differential equations and graph neural networks, unlocking recent advances in continuous--depth learning for non--trivial topologies.
\paragraph{\fcircle[fill=deblue]{3pt} Blending graphs and differential equations}
We introduce the system--theoretic model class of \textit{neural graph differential equations} (Neural GDEs), defined as ODEs parametrized by GNNs. Neural GDEs are designed to inherit the ability to impose relational inductive biases of GNNs while retaining the dynamical system perspective of continuous--depth models. A complete model taxonomy is carefully laid out with the primary objective of ensuring compatibility with modern GNN variants. Neural GDEs offer a grounded approach for the embedding of numerical schemes inside the forward pass of GNNs, in both the deterministic as well as the stochastic case.

\paragraph{\fcircle[fill=wine]{3pt} Dynamic graphs}
Additional Neural GDE variants are developed to tackle the spatio--temporal setting of dynamic graphs. In particular, we formalize general Neural Hybrid GDE models as \textit{hybrid dynamical systems} \citep{van2000introduction,goebel2008}. Here, the structure--dependent vector field learned by Neural GDEs offers a data--driven approach to the modeling of dynamical networked systems \citep{lu2005time,andreasson2014distributed}, particularly when the governing equations are highly nonlinear and therefore challenging to approach with analytical methods. Neural GDEs can adapt the prediction horizon by adjusting the integration interval of the differential equation, allowing the model to track evolution of the underlying system from irregular observations. The evaluation protocol for Neural GDEs spans several application domains, including traffic forecasting and prediction in biological networks.
\section{Neural GDEs \fcircle[fill=deblue]{3pt}}
We begin by introducing the general formulation. We then provide a taxonomy for Neural GDE models, distinguishing them into \textit{static} and \textit{spatio--temporal} variants. 
\subsection{General Framework}
Without any loss of generality, the inter--layer dynamics of a residual graph neural network (GNN) may be represented in the form:
\begin{equation}\label{resgnn}
    \left\{
        \begin{aligned}
            Z_{k+1} &= Z_k + f_{\cG}^k\lc k, Z_k, \theta_k\rc\\
            Z_0 &= \ell^x_{\cG} (X)\\
            \hat{Y}_k &= \ell^y_{\cG}(Z_k)
        \end{aligned}
    \right.,~~k\in\Nat,
\end{equation}
with hidden state $Z_k\in\R^{n\times \nZ}$, node features $X\in\R^{n\times \nX}$ and output $\hat{Y}\in\R^{n\times \nY}$. $f^k_{\cG}$ are generally matrix--valued nonlinear functions conditioned on graph $\cG$, $\theta_k\in\R^{\nT}$ is the tensor of trainable parameters of the $k$-th layer and $\ell_{\cG}^x,~\ell_{\cG}^y$ represent feature embedding and output layers, respectively. Note that the explicit dependence on $k$ of the dynamics is justified in some graph architectures, such as diffusion graph convolutions \citep{atwood2016diffusion}. 

A \textit{neural graph differential equation} (Neural GDE) is constructed as the continuous--depth limit of \eqref{resgnn}, defined as the nonlinear affine dynamical system:
\begin{equation}\label{eq:GDE}
    \left\{
        \begin{aligned}
            \dot Z_t &= f_{\cG}\lc t, Z_t, \theta_t\rc \\
            Z_0 &= \ell_{\cG}^x (X)\\
            \hat{Y}_t &= \ell_{\cG}^y (Z_t)
        \end{aligned}
    \right.,~~s\in\cS\subset\R,
\end{equation}
where $f_{\cG}:\cT\times\R^{n\times\nZ}\times\R^\nT\rightarrow\R^{n\times\nZ}$ is a depth--varying vector field defined on graph $\cG$ and $\ell_{\cG}^x:\R^{n\times n_x}\rightarrow \R^{n\times n_z},~\ell^y_{\cG}:\R^{n\times n_z}\rightarrow \R^{n\times n_y}$ are two affine linear mappings. Depending on the choice of input transformation $\ell_{\cG}^x$, different node feature augmentation techniques can be introduced \citep{dupont2019augmented,massaroli2020dissecting} to reduce stiffness of learned vector fields.

\paragraph{Well--posedness} Let $\cT:=[0,1]$. Under mild conditions on $f_{\cG}$, namely Lipsichitz continuity with respect to $Z$ and uniform continuity with respect to $t$,  for each initial condition (GDE embedded input) $Z_0=\ell_{\cG}^x (X)$, the matrix--valued ODE in \eqref{eq:GDE} admits a unique solution $Z_t$ defined in the whole $\cS$. Thus there is a mapping $\Phi$ from $\R^{n\times \nZ}$ to the space of absolutely continuous functions $\cT\to\R^{n\times \nZ}$ such that $Z_t := \Phi_t(X)$ satisfies the ODE in \eqref{eq:GDE}. 
Symbolically, the output of the Neural GDE is obtained by the following
\begin{equation*}
    \hat{Y}_t = \ell_{\cG}^y\Big(\ell^x_{\cG}(X) + \int_0^t f_{\cG}(s ,Z_s),\theta)\dd s\Big).
\end{equation*}
\vspace{-5mm}

\subsection{Neural GDEs on Static Graphs}
Based on graph spectral theory \citep{shuman2013emerging,sandryhaila2013discrete}, the residual version of \textit{graph convolution network} (GCN) \citep{kipf2016semi} layers is obtained by setting:
\begin{equation}\label{eq:gcn}
     f_{\cG}^k(Z_k, \theta_k) :=L_{\cG} Z_k W_k
\end{equation}
in \eqref{resgnn}, where $L_{\cG}\in\R^{n\times n}$ is the graph \textit{Laplacian}, $\theta_k:=\vect({W_k})$ and $\sigma$. 
The general formulation of the continuous GCNs counterpart, \textit{neural graph convolution differential equations} (Neural GCDEs) is similarly defined by letting the vector field $f_{\cG}$ be a multilayer convolution, i.e.
\begin{equation}\label{eq:gde}
    \dot Z_t = f_{\cG}(Z_t, \theta) := f^N_{\cG}\circ\sigma\circ f^{N-1}_{\cG}\circ\sigma\circ\cdots\circ f^1_{\cG}
\end{equation}
with $\sigma$ being a nonlinear activation function though to be acting element--wise and $\theta:=\vect(W_1, \dots, W_N)\in\R^{n_z^2}$. Note that the Laplacian $L_{\cG}$ can be computed in different ways, see e.g. \citep{bruna2013spectral,defferrard2016convolutional,levie2018cayleynets, zhuang2018dual}. Alternatively, diffusion--type convolution layers \citep{li2017diffusion} can be introduced. We note that expressivity of the model is improved via letting the parameters be \textit{time--varying}  i.e. $f_{\cG}(Z_t, \theta_t) := L_{\cG}Z_tW_t$, $\theta_t :=  \vect(W_t)$ where $\theta_t$ is parametrized by spectral or time discretizations \citep{massaroli2020dissecting}.
\paragraph{Additional continuous--time variants} We include additional derivations of continuous--time counterparts of common static GNN models such as \textit{graph attention networks} (GATs) \citep{velivckovic2017graph} and general message passing GNNs in the Appendix. We note that due to the purely algebraic nature of common operations in geometric models such as attention operators \citep{vaswani2017attention}, Neural GDEs are compatible with the vast majority of GNN architectures.
\section{Neural GDEs on Dynamic Graphs \fcircle[fill=wine]{3pt}}
Common use cases for GNNs involve prediction in dynamic graphs, which introduce additional challenges. We discuss how Neural GDE models can be extended to address these scenarios, leveraging tools from \textit{hybrid dynamical system} theory \citep{van2000introduction} to derive a Neural Hybrid GDE formulation.

Here, Neural GDEs represent a natural model class for autoregressive modeling of sequences of graphs $\{\cG_t\}$ and seamlessly link to dynamical network theory. This line of reasoning naturally leads to an extension of classical spatio--temporal architectures in the form of \textit{hybrid dynamical systems} \citep{van2000introduction,goebel2008}, i.e., systems characterized by interacting continuous and discrete--time dynamics.
\vspace{-2mm}
\paragraph{Notation:}
Let  $(\cK, >)$, $(\cT_e, >)$ be linearly ordered sets; namely,  $\cK\subset\Nat\setminus\{0\}$ and $\cT_e$ is a set of time instants, $\cT_e:=\{t_k\}_{k\in\cK}$. We suppose to be given a \textit{state--graph data stream} which is a sequence in the form $\left\{\left(X_t,\cG_t\right)\right\}_{t\in\cT_e}$. 
Let us also define a \textit{hybrid time domain} as the set $\I:= \bigcup_{k\in\cK}\lc[t_k, t_{k+1}],k\rc$ and a \textit{hybrid arc} on $\I$ as a function $\Phi$ such that for each $k\in\cK$, $t\mapsto\Phi(t,k)$ is absolutely continuous in $\{t:(t,j)\in\dom\Phi\}$. Our aim is to build a continuous model predicting, at each $t_k\in\cT_e$, the value of $X_{t_{k+1}}$, given $\left(X_t,\cG_t\right)$.

\subsection{Neural Hybrid GDEs}\label{sec:3.1}
The core idea is to have a Neural GDE smoothly steering the latent node features between two time instants and then apply some discrete transition operator, resulting in a ``jump'' of state $Z$ which is then processed by an output layer. Solutions of the proposed continuous spatio--temporal model are therefore hybrid arcs.  

The general formulation of a Neural Hybrid GDE model can be symbolically represented by:
\begin{equation}\label{eq:hybrid}
  \left\{
        \begin{matrix*}[l]
            \dot{Z}_t &= f_{\cG_{t_k}}(Z_t, \theta) & t\in[t_{k-1}, t_{k}] \\[3pt]

            Z^+_t &= \ell^j_{\cG_{{t_k}}}(Z_t, X_{t}) & t = t_{k}\\[3pt]
            \hat{Y}_t &= \ell^y_{\cG_{t_k}}(Z_t) 
        \end{matrix*}
    \right.k\in\K,
\end{equation} 
where $f_{\cG}, \ell^j_{\cG}, \ell_{\cG}^y$ are GNN--like operators or general neural network layers and $Z^+_t$ represents the value of $Z_t$ after the discrete transition. The evolution of system (\ref{eq:hybrid}) is indeed a sequence of hybrid arcs defined on a hybrid time domain. Compared to standard recurrent models which are only equipped with discrete jumps, system (\ref{eq:hybrid}) incorporates a continuous flow of latent node features $Z_t$ between jumps. This feature of Hybrid Neural GDEs allows them to track the evolution of dynamical systems from observations with irregular time steps. In the experiments we consider $\ell^j_{\cG}$ to be a GRU cell \citep{cho2014learning}, obtaining \textit{neural graph convolution differential equation--GRU} (GCDE--GRU).

\paragraph{Sequential adjoint for Hybrid GDEs}
Continuous--depth models, including Neural ODEs and SDEs, can be trained using adjoint sensitivity methods \citep{pontryagin1962mathematical,chen2018neural,li2020scalable}. Care must be taken in the case of sequence models such as Neural Hybrid GDEs which often admit losses dependent on solution values at various timestamps, rather than considering terminal states exclusively. In particular, we may consider loss functions of the form
\[
    L_{\theta}(X_{t_{0:N}}, Z_{t_{0:N}}) = \sum_{k\in \cK} c_\theta(X_{t_k}, Z_{t_k}).
\]
In such a case,  back--propagated gradient can be computed with an extension of classic adjoint techniques \citep{pontryagin1962mathematical}
\[
    \frac{\dd L_\theta}{d\theta} = \frac{\partial L_\theta}{\partial \theta}-\int_{\cT} \left<\lambda, \frac{\partial f_{\cG_t}(t, \Phi_{t}(Z_0),\theta)}{\partial\theta}\right>\dd t
\]
where the Lagrange multiplier $\lambda_t:\I^- \rightarrow\R^{n\times n_x}$ is, however, a hybrid arc on the reversed (backward) hybrid time domain $\I^-$ satysfying the hybrid inclusion
\[
    \begin{aligned}
        \dot \lambda_t &\in F(t, Z_t, \lambda_t)&& ~~~~~t \in \cT\\
        \lambda^+_t & \in G(X_t, Z_t, \lambda_t)&& ~~~~t\in \cT_e\\
        \lambda_{t_K} &= \frac{\partial c_\theta}{\partial Z_{t_K}} 
    \end{aligned}
\]
with $t_K=\sup\cT_e$ and $F,~G$ are set--valued mappings $F:\cT\times\R^{n\times n_z}\times\R^{n\times n_z}\rightrightarrows\R^{n\times n_z}$,~$G:\R^{n\times n_x}\times\R^{n\times n_z}\times\R^{n\times n_z}\rightrightarrows\R^{n\times n_z}$,~defined as
\[ 
    \begin{aligned}
        F(t, Z_t, \lambda_t):=\left\{F_k:t\in[t_k,t_{k+1}]\Rightarrow F_k=-\frac{\partial f_{\cG_{t_k}}}{\partial Z}\lambda_t\right\}\\
        G(X_t, Z_t, \lambda_t):=\left\{G_k:t=t_k\Rightarrow G_k= \lambda_t + \frac{\partial c_\theta}{\partial Z_t}\right\}.
    \end{aligned}
\]
\begin{figure*}
    \centering
    \includegraphics[width=1\linewidth]{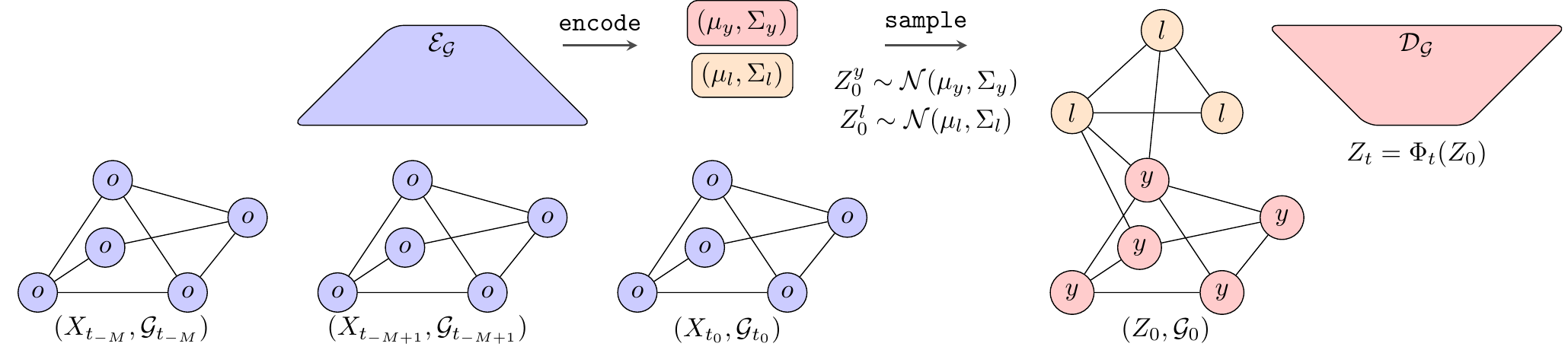}
    \caption{\footnotesize \textit{Latent Neural GDE} model. Data observations are encoded into a compact representation, required to sample from approximate posteriors on observable and latent node features, forming an \textit{augmented} graph. Following this procedure, the dynamics are unrolled by the Neural GDE decoder $\D_{\G}$. We consider deterministic as well as stochastic decoders in the form of Neural GSDEs.}
    \label{fig:latentschematic}
\end{figure*}

\subsection{Latent Neural GDEs}
The Neural GDE variants introduced so far can be combined into a latent variable model for dynamic graphs. In particular, we might be interested in predicting $(X_{t_1}, \cG_{t_1}), \dots, (X_{t_N}, \cG_{t_N})$ given past observations $(X_{t_{-M}}, \cG_{t_{-M}}), \dots, (X_{t_0}, \cG_{t_0})$. To this end, we introduce \textit{Latent Neural GDEs} as {\color{blue!50}encoder}--{\color{wine}decoder} models in the form
\begin{equation*}
    \begin{aligned}
        ~~&E = {\color{blue!40}\E_{\cG}}(\G_{t_{-M:0}}, X_{t_{-M:0}})~~~~~~~~~~~~~~~~~{\color{blue!40}\text{Graph Encoder}}\\
        ~~&Z_0 \sim q_{\cG}(Z_0|E)~~~~~~~~~~~~~~~~~~~~~~~~~~~~{\color{black!80}\text{Reparametrization}}\\
        ~~&{Z}_{t} = \Phi_t(Z_0)~~~~~~~~~~~~~~~~~~~~~~~~~~~~~~~~~~~~~{\color{wine}\text{Graph Decoder}}\\
        ~~&\hat{Y}_t = Z_t^y
    \end{aligned}
\end{equation*}
where the decoded output $\hat{Y}_t$ in a time--domain $\mathcal{T}$ is obtained via the solution of Neural GDE. Latent Neural GDEs are designed for reconstruction or extrapolation tasks involving dynamic graphs, where the underlying data--generating process is known to be a differential equation. 

Rather than constraining the evolution of decoded node features to a latent space, requiring readout layers to map back to data--space, we construct an \textit{augmented} graph with latent nodes $Z^l$ and output--space nodes $Z^y$. This formulation retains the flexibility of a full latent model while allowing for the embedding of stricter inductive biases on the nature of the latent variables and their effects, as shown experimentally on genetic regulatory networks. Figure \ref{fig:latentschematic} depicts an example instance of Latent Neural GDEs, where the approximate posterior on $Z^l$ and $Z^y$ is defined as a multivariate Gaussian.

Latent Neural GDEs are trained via maximum likelihood. The optimization problem can be cast as the maximization of an evidence lower bound (${\tt ELBO}$):  

\begin{equation*}\label{elbo}
    \begin{aligned}
        {\tt ELBO} &:= \mathbb{E}_{Z_0\sim q_{\cG}}\Big[\sum_{k=1}^N\log p(\hat{Y}_{t_k}) - {\tt KL}(q_{\cG}||\mathcal{N}(0,\mathbb{I}))\Big] 
    \end{aligned}
\end{equation*}
with an observation--space density defined as $p(\hat{Y}_{t_k}) = \mathcal{N}(Y_{t_k}, \Sigma_{t_k})$ and a covariance $\Sigma_{t_k}$ hyperparameter.
\paragraph{Embedding stochasticity into Neural GDEs}

Recent work \citep{li2020scalable,peluchetti2020infinitely,massaroli2021learning} develops extensions of continuous models to stochastic differential equations \citep{kunita1997stochastic, oksendal2003stochastic} for static tasks or optimal control. These results carry over to the Neural GDEs framework; an example application is to consider \textit{stochastic} Latent Neural GDE decoders in order to capture inherent stochasticity in the samples. Here, given a multi--dimensional Browian motion  $B_t$, we define and train \textit{Neural graph stochastic differential equations} (Neural GSDEs) of the form
\[ \dd Z_t = f_{\cG}(t, Z_t, \theta)\dd t + g_{\cG}(t, Z_t, \theta)\circ \dd B_t \] 
where we adopt a \textit{Stratonovich} SDE formulation. 
\section{Experiments \fcircle[fill=olive]{3pt}}
We evaluate Neural GDEs on a suite of different tasks. The experiments and their primary objectives are summarized below:
\begin{itemize}[leftmargin=0.2in]
    \item Trajectory extrapolation task on a synthetic multi--agent dynamical system. We compare Neural ODEs and Neural GDEs, providing in addition to the comparison a motivating example for the introduction of additional biases inside GDEs in the form of second--order models \citep{yildiz2019ode,massaroli2020dissecting,norcliffe2020second}. 
    \item Traffic forecasting on an undersampled version of PeMS \citep{yu2018spatio} dataset. We measure the performance improvement obtained by a correct inductive bias on continuous dynamics and robustness to irregular timestamps.
    \item Flux prediction in genetic regulatory networks such as Elowitz-Leibler repressilator circuits  \citep{Elowitz2000}. We investigate Latent Neural GDEs for prediction in biological networks with stochastic dynamics, where prior knowledge on graph structure linking latent and observable nodes plays a key role.
 \end{itemize}  

\subsection{\fcircle[fill=wine]{3pt} Multi--Agent Trajectory Extrapolation}
\paragraph{Experimental setup}
We evaluate GDEs and a collection of deep learning baselines on the task of extrapolating the dynamical behavior of a synthetic, non--conservative mechanical multi--particle system. Particles interact within a certain radius with a viscoelastic force. Outside the mutual interactions, captured by a time--varying adjacency matrix $A_t$, the particles would follow a periodic motion, gradually losing energy due to viscous friction. The adjacency matrix ${A}_t$ is computed along the trajectory as:
\[
    A_t^{(ij)} = 
    \left\{\begin{matrix*}[l]
        1 & 2\|x_i(t)-x_j(t)\|\leq r\\
        0 & \text{otherwise}
    \end{matrix*}\right.~,
\]
where $\x_i(t)$ is the position of node $i$ at time $t$. Therefore, ${A}_t$ results to be symmetric, ${A}_t = {A}_t^\top$ and yields an undirected graph. The dataset is collected by integrating the system for $T = 5s$ with a fixed step--size of $dt = 1.95\cdot10^{-3}$ and is split evenly into a training and test set. We consider $10$ particle systems. An example trajectory is shown in Appendix C. All models are optimized to minimize mean--squared--error (MSE) of 1--step predictions using Adam \citep{kingma2014adam} with constant learning rate $0.01$. 
We measure test \textit{mean average percentage error} (MAPE) of model predictions in different extrapolation regimes. \textit{Extrapolation steps} denotes the number of predictions each model $\Phi$ has to perform without access to the nominal trajectory. This is achieved by recursively letting inputs at time $t$ be model predictions at time $t - \Delta t$ i.e $\hat{Y}_{t+\Delta t} = \phi (\hat{Y}_{t})$ for a certain number of extrapolation steps, after which the model is fed the actual nominal state $X$ and the cycle is repeated until the end of the test trajectory. For a robust comparison, we report mean and standard deviation across 10 seeded training and evaluation runs. Additional experimental details, including the analytical formulation of the dynamical system, are provided as supplementary material. 
\paragraph{Models}
As the vector field depends only on the state of the system, available in full during training, the baselines do not include recurrent modules. We consider the following models:
\begin{itemize}[leftmargin=0.2in]
    \item A 3--layer fully-connected neural network, referred to as \textit{Static}. No assumption on the dynamics
    \item A vanilla Neural ODE with the vector field parametrized by the same architecture as \textit{Static}. ODE assumption on the dynamics.
    \item A 3--layer convolution Neural GDE, Neural GCDE. Dynamics assumed to be determined by a blend of graphs and ODEs
    \item A 3--layer, second--order \citep{yildiz2019ode,massaroli2020dissecting,norcliffe2020second} Neural GCDE and referred to as \textit{Neural GCDE-II}. 
\end{itemize}
A grid hyperparameter search on number of layers, ODE solver tolerances and learning rate is performed to optimize \textit{Static} and Neural ODEs. We use the same hyperparameters for Neural GDEs. We used the {\tt torchdyn} \cite{poli2020torchdyn} library for Neural ODE baselines.

\begin{wrapfigure}[12]{r}{0.43\textwidth}
    \centering
    \vspace{-3mm}
    \includegraphics[width=1\linewidth]{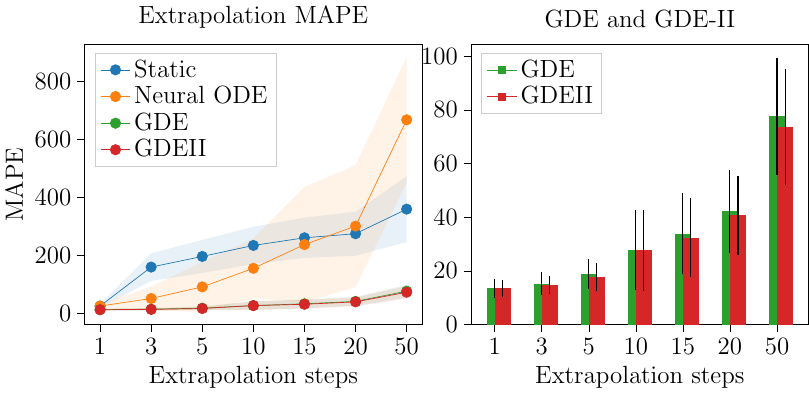}
    \vspace{-6mm}
    \caption{\footnotesize Test extrapolation MAPE averaged across 10 experiments. Shaded area and error bars indicate 1--standard deviation intervals. }
    \label{fig:mape}
\end{wrapfigure}

 \begin{table*}[t]
 \small
\centering
\setlength\tabcolsep{3pt}
\begin{tabular}{lrrrrrrrr}
\toprule
Model & MAPE$_{30\%}$ & RMSE$_{30\%}$ & MAPE$_{70\%}$ & RMSE$_{70\%}$ &  MAPE$_{100\%}$ & RMSE$_{100\%}$\\
\midrule
GRU & $27.14 \pm 0.45$ & $13.25 \pm 0.11$ & $27.24 \pm 0.19$ & $13.28 \pm 0.05$ & $27.20 \pm 0.00$ & $13.29 \pm 0.00$\\
GCGRU & $23.60 \pm 0.38$ & $11.97 \pm 0.06$ & $21.33 \pm 0.16$ & $11.20 \pm 0.04$ &$20.92 \pm 0.00$ & $10.87 \pm 0.00$\\
GCDE-GRU & $\mathbf{22.95} \pm 0.37$ & $\mathbf{11.67} \pm 0.10$ & $\mathbf{20.94} \pm 0.14$ & $\mathbf{10.95} \pm 0.04$ & $\mathbf{20.46} \pm 0.00$ & $\mathbf{10.76} \pm 0.00$\\
\bottomrule
\end{tabular}
\caption{\footnotesize Forecasting test results across 20 runs (mean and standard dev.). MAPE$_i$ indicates an $i\%$ test sampling undersampling strategy, i.e $i\%$ of the time series measurements are randomly selected and kept for training.}
\label{tab:traffic_preds}
\end{table*}
\paragraph{Results}
Figure \ref{fig:mape} shows the growth rate of test MAPE error as the number of extrapolation steps is increased. \textit{Static} fails to extrapolate beyond the 1--step setting seen during training. Neural ODEs overfit spurious particle interaction terms and their error rapidly grows as the number of extrapolation steps is increased. Neural GCDEs, on the other hand, are able to effectively leverage relational information to track the system: we provide complete visualization of extrapolation trajectory comparisons in the Appendix. Lastly, Neural GCDE-IIs outperform first--order Neural GCDEs as their structure inherently possesses crucial information about the relative relationship of positions and velocities, accurate with respect to the observed dynamical system. 
\subsection{\fcircle[fill=wine]{3pt} Traffic Forecasting}
\paragraph{Experimental setup}

We evaluate the effectiveness of Neural Hybrid GDE models on forecasting tasks by performing a series of experiments on the established PeMS traffic dataset. We follow the setup of \citep{yu2018spatio} in which a subsampled version of PeMS, PeMS7(M), is obtained via selection of 228 sensor stations and aggregation of their historical speed data into regular 5 minute frequency time series. We construct the adjacency matrix $A$ by thresholding the Euclidean distance between observation stations i.e. when two stations are closer than the threshold distance, an edge between them is included. The threshold is set to the 40$^{\text{th}}$ percentile of the station distances distribution. To simulate a challenging environment with missing data and irregular timestamps, we undersample the time series by performing independent Bernoulli trials on each data point. Results for 3 increasingly challenging experimental setups are provided: undersampling with $0\%$, $30\%$, $50\%$ and $70\%$ of removal. In order to provide a robust evaluation of performance in regimes with irregular data, the testing is repeated $20$ times per model, each with a different undersampled version of the test dataset. We collect \textit{root mean square error} (RMSE) and MAPE. More details about the chosen metrics and data are included as supplementary material.
\begin{figure}[t]
    \centering
    \input{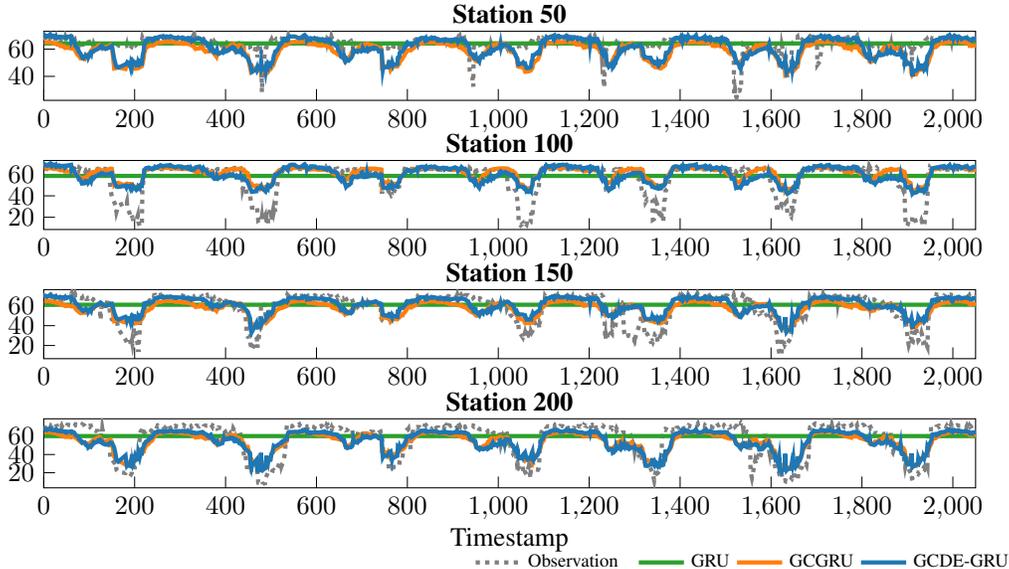}
    \vspace{-3mm}
    \caption{\footnotesize Traffic data prediction results of 50\% undersampling. GCDE--GRUs are able to evolve the latents between timestamps and provide a more accurate fit.}
    \label{fig:traffic_preds}
\end{figure}
\paragraph{Models}
In order to measure performance gains obtained by Neural GDEs in settings with data generated by continuous time systems, we employ a GCDE--GRU--dpr5 as well as its discrete counterpart GCGRU \citep{zhao2018deep}. To contextualize the effectiveness of introducing graph representations, we include the performance of GRUs since they do not directly utilize structural information of the system in predicting outputs. Apart from GCDE--GRU, both baselines have no innate mechanism for handling timestamp information. For a fair comparison, we include timestamp differences between consecutive samples and \textit{sine--encoded} \citep{petnehazi2019recurrent} absolute time information as additional features. All models receive an input sequence of $5$ graphs to perform the prediction.
\paragraph{Results}
Non--constant differences between timestamps result in a challenging forecasting task for a single model since the average prediction horizon changes drastically over the course of training and testing. 
Traffic systems are intrinsically dynamic and continuous in nature and, therefore, a model able to track continuous underlying dynamics is expected to offer improved performance.
Since GCDE-GRUs and GCGRUs are designed to match in structure we can measure this performance increase from the results shown in Table \ref{tab:traffic_preds}. GCDE--GRUs outperform GCGRUs and GRUs in all undersampling regimes. Additional details and visualizations are included in Appendix C.
\subsection{\fcircle[fill=wine]{3pt} Repressilator Reconstruction}

\begin{wrapfigure}[16]{r}{0.41\textwidth}
\vspace{-16mm}
    \centering
    \begin{tikzpicture}[thick,scale=0.7, every node/.style={scale=0.7}]

\usetikzlibrary{calc,patterns,positioning,decorations.pathmorphing}

\definecolor{mygreen}{HTML}{0c9463}
\definecolor{mymauve}{rgb}{0.58,0,0.82}
\definecolor{echodrk}{HTML}{0099cc}
\definecolor{darkgrey}{RGB}{61,72,73}

\node[circle, fill=mymauve, draw=mymauve, inner sep=0.1em, thick] (h1) {\textcolor{white}{$h_0$}};
\node[circle, draw, densely dotted, mymauve, thick, inner sep=1em] at (h1) {};


\node[circle, draw=blue!70!white, opacity=.7, fill=blue!70!white, opacity=.7, inner sep=0.1em, thick] (001) at ([shift=({90:6 em})]h1) {\textcolor{white}{${\tt LaCI} \atop \text{mRNA}$}};
\node[circle, draw=blue!70!white, opacity=.7, fill=blue!70!white, opacity=.7, inner sep=0.1em, thick] (002) at ([shift=({210:6 em})]h1) {\textcolor{white}{${\tt TetR} \atop \text{mRNA}$}};
\node[circle, draw=blue!70!white, opacity=.7, fill=blue!70!white, opacity=.7, inner sep=0.1em, thick] (003) at ([shift=({330:6 em})]h1) {\textcolor{white}{${\tt cI} \atop \text{mRNA}$}};

\node[circle, draw=mygreen, fill=mygreen, inner sep=0.1em, thick] (011) at ([shift=({90:3 em})]001) {\textcolor{white}{$r$}};
\node[circle, draw=mygreen, fill=mygreen, inner sep=0.1em, thick] (012) at ([shift=({210:3 em})]002) {\textcolor{white}{$r$}};
\node[circle, draw=mygreen, fill=mygreen, inner sep=0.1em, thick] (013) at ([shift=({330:3 em})]003) {\textcolor{white}{$r$}};

\draw[-stealth, thick, gray] (001) -- (011);
\draw[-stealth, thick, gray] (002) -- (012);
\draw[-stealth, thick, gray] (003) -- (013);

\node[circle, draw=blue!70!white, opacity=.7, fill=blue!70!white, opacity=.7, inner sep=0.1em, thick] (021) at ([shift=({90+45:3 em})]011) {\textcolor{white}{${\tt LaCI} \atop \text{prot}$}};
\node[circle, draw=blue!70!white, opacity=.7, fill=blue!70!white, opacity=.7, inner sep=0.1em, thick] (022) at ([shift=({210+45:3 em})]012) {\textcolor{white}{${\tt TetR} \atop \text{prot}$}};
\node[circle, draw=blue!70!white, opacity=.7, fill=blue!70!white, opacity=.7, inner sep=0.1em, thick] (023) at ([shift=({330+45:3 em})]013) {\textcolor{white}{${\tt cI} \atop \text{prot}$}};
\draw[-stealth, thick, gray] (011) edge[bend left=10, decoration={}, decorate] (021);
\draw[-stealth, thick, gray] (012) edge[bend left=10, decoration={}, decorate] (022);
\draw[-stealth, thick, gray] (013) edge[bend left=10, decoration={}, decorate] (023);

\node[circle, draw=mymauve, fill=mymauve, inner sep=0.1em, thick] (031) at ([shift=({90-45:3 em})]011) {\textcolor{white}{$h_1$}};
\node[circle, draw=mymauve, fill=mymauve, inner sep=0.1em, thick] (032) at ([shift=({210-45:3 em})]012) {\textcolor{white}{$h_2$}};
\node[circle, draw=mymauve, fill=mymauve, inner sep=0.1em, thick] (033) at ([shift=({330-45:3 em})]013) {\textcolor{white}{$h_3$}};
\draw[stealth-, thick, gray] (011) edge[bend left=-10, decoration={}, decorate] (031);
\draw[stealth-, thick, gray] (012) edge[bend left=-10, decoration={}, decorate] (032);
\draw[stealth-, thick, gray] (013) edge[bend left=-10, decoration={}, decorate] (033);

\node[circle, draw=mygreen, fill=mygreen, inner sep=0.1em, thick] (041) at ([shift=({90:4 em})]011) {\textcolor{white}{$r$}};
\node[circle, draw=mygreen, fill=mygreen, inner sep=0.1em, thick] (042) at ([shift=({210:4 em})]012) {\textcolor{white}{$r$}};
\node[circle, draw=mygreen, fill=mygreen, inner sep=0.1em, thick] (043) at ([shift=({330:4 em})]013) {\textcolor{white}{$r$}};
\draw[-stealth, thick, gray] (021) edge[bend left=10, decoration={}, decorate] (041);
\draw[-stealth, thick, gray] (022) edge[bend left=10, decoration={}, decorate] (042);
\draw[-stealth, thick, gray] (023) edge[bend left=10, decoration={}, decorate] (043);
\draw[-stealth, thick, gray] (041) edge[bend left=10, decoration={}, decorate] (031);
\draw[-stealth, thick, gray] (042) edge[bend left=10, decoration={}, decorate] (032);
\draw[-stealth, thick, gray] (043) edge[bend left=10, decoration={}, decorate] (033);

\node[circle, draw=mygreen, fill=mygreen, inner sep=0.1em, thick] (111) at ([shift=({210:4 em})]001) {\textcolor{white}{$r$}};
\node[circle, draw=mygreen, fill=mygreen, inner sep=0.1em, thick] (112) at ([shift=({330:4 em})]001) {\textcolor{white}{$r$}};
\draw[stealth-, thick, darkgray] (h1) edge[bend left=18, decoration={}, decorate] (111);
\draw[stealth-, thick, darkgray] (111) edge[bend left=18, decoration={}, decorate] (001);
\draw[-stealth, thick, darkgray] (h1) edge[bend left=-18, decoration={}, decorate] (112);
\draw[-stealth, thick, darkgray] (112) edge[bend left=-18, decoration={}, decorate] (001);

\node[circle, draw=mygreen, fill=mygreen, inner sep=0.1em, thick] (211) at ([shift=({90:4 em})]002) {\textcolor{white}{$r$}};
\node[circle, draw=mygreen, fill=mygreen, inner sep=0.1em, thick] (212) at ([shift=({330:4 em})]002) {\textcolor{white}{$r$}};
\draw[-stealth, thick, darkgray] (h1) edge[bend left=-18, decoration={}, decorate] (211);
\draw[-stealth, thick, darkgray] (211) edge[bend left=-18, decoration={}, decorate] (002);
\draw[stealth-, thick, darkgray] (h1) edge[bend left=18, decoration={}, decorate] (212);
\draw[stealth-, thick, darkgray] (212) edge[bend left=18, decoration={}, decorate] (002);

\node[circle, draw=mygreen, fill=mygreen, inner sep=0.1em, thick] (311) at ([shift=({210:4 em})]003) {\textcolor{white}{$r$}};
\node[circle, draw=mygreen, fill=mygreen, inner sep=0.1em, thick] (312) at ([shift=({90:4 em})]003) {\textcolor{white}{$r$}};
\draw[-stealth, thick, darkgray] (h1) edge[bend left=-18, decoration={}, decorate] (311);
\draw[-stealth, thick, darkgray] (311) edge[bend left=-18, decoration={}, decorate] (003);
\draw[stealth-, thick, darkgray] (h1) edge[bend left=18, decoration={}, decorate] (312);
\draw[stealth-, thick, darkgray] (312) edge[bend left=18, decoration={}, decorate] (003);

\draw[-stealth, thick, darkgrey] (021) edge[bend left=-25, decoration={}, decorate] (211);
\draw[-stealth, thick, darkgrey] (022) edge[bend left=-25, decoration={}, decorate] (311);
\draw[-stealth, thick, darkgrey] (023) edge[bend left=-25, decoration={}, decorate] (112);

\end{tikzpicture}
    \vspace{-6mm}
    \caption{\footnotesize A bipartite graph representation of a metabolic flux schematic for the Elowitz-Leibler repressilator, where $h$ is degradation, and $r$ is reaction.}
    \label{fig:repressilator}
\end{wrapfigure}
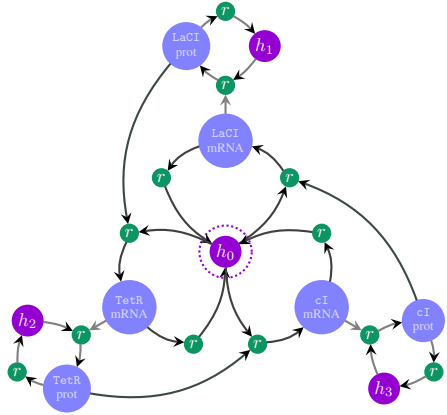

\paragraph{Experimental setup}
We investigate prediction of genetic regulatory network dynamics as a showcase application of Latent Neural GDEs. In particular, we consider biological metabolic fluxes of an Elowitz-Leibler repressilator circuit~\citep{Elowitz2000}. Repressilator circuits are a common example of feedback mechanism used to maintain homeostasis in biological systems. The circuit is modeled as a bipartite graph where one set of nodes is comprised of reactions and a second one of biochemical species (mRNA and protein), as shown in Figure~\ref{fig:repressilator}. 
The repressilator feedback cycle is structured such that each protein suppresses the expression of mRNA for the next protein in the cycle, yielding oscillatory behaviour.
Accurate genome-scale models of dynamic metabolic flux are currently intractable, largely due to scaling limitations or insufficient prediction accuracy when compared with \textit{in vitro} and \textit{in vivo} data. By example, state--of--the--art genome-scale dynamic flux simulations currently require model reduction to core metabolism~\citep{Masid2020}. In order to demonstrate the effectiveness of the Neural GDE framework in fitting stochastic systems and allowing for interpretability of underlying mechanisms, we generate a training dataset of ten trajectories by symbolic integration via the $\tau$--leaping method \citep{gillespie2007, padgett2016tau} over a time span of $300$ seconds. During training, we split each trajectory into halves and task the model with reconstruction of the last $150$ seconds during the decoding phase, conditioned on the first half.
\begin{wrapfigure}[15]{r}{0.55\textwidth}
    \centering
    \vspace{-7mm}
    \includegraphics[width=\linewidth]{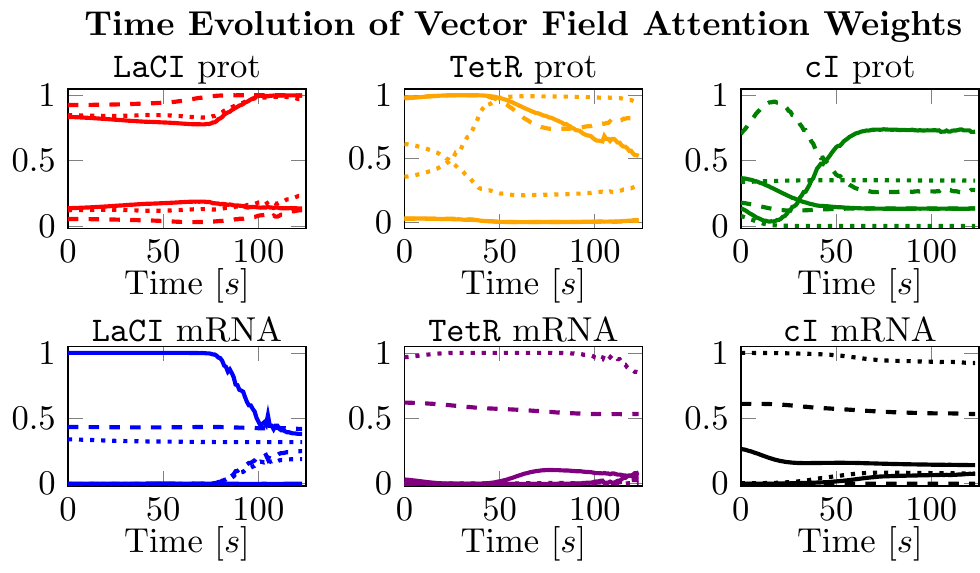}
    \vspace{-7mm}
    \caption{\footnotesize Evolution of GAT \cite{velivckovic2017graph} attention weights of drift networks $f_{\cG}$ in the Neural GSDE decoder. The linestyles indicate different edge weights (\textit{incoming} and \textit{outgoing}) for the three edges connected to each protein and mRNA. }
    \label{fig:attentionbois}
\end{wrapfigure}
\paragraph{Models}
We assess modeling capabilities of Latent Neural GDEs applied to biological networks, with a focus on interpretability. The particular graph structure of the system shown in Figure~\ref{fig:repressilator} lends itself to a formulation where protein and mRNA dynamics can be grouped in $Z^y$, whereas reaction nodes constitute the set of latent nodes $Z^l$. This in turn allows the decoder to utilize prior knowledge on the edges connecting the two sets of nodes. We report the full adjacency matrix in Appendix B. 

The Latent Neural GDE is equipped with an encoder comprised of 2--layers of \textit{temporal convolutions} (TCNs). To model stochasticity and provide uncertainty estimates in predictions, the architecture leverages \textit{neural graph stochastic differential equation} (Neural GSDE) during decoding steps.
The drift $f_{\cG}$ and diffusion $g_{\cG}$ networks of the Neural GSDEs follow an equivalent 3--layer GNN design: [GCN, GAT \citep{velivckovic2017graph}, GCN] with hidden dimension $3$ and hyperbolic--tangent activations.
\begin{figure*}[t]
    \centering
    \includegraphics[width=1\linewidth]{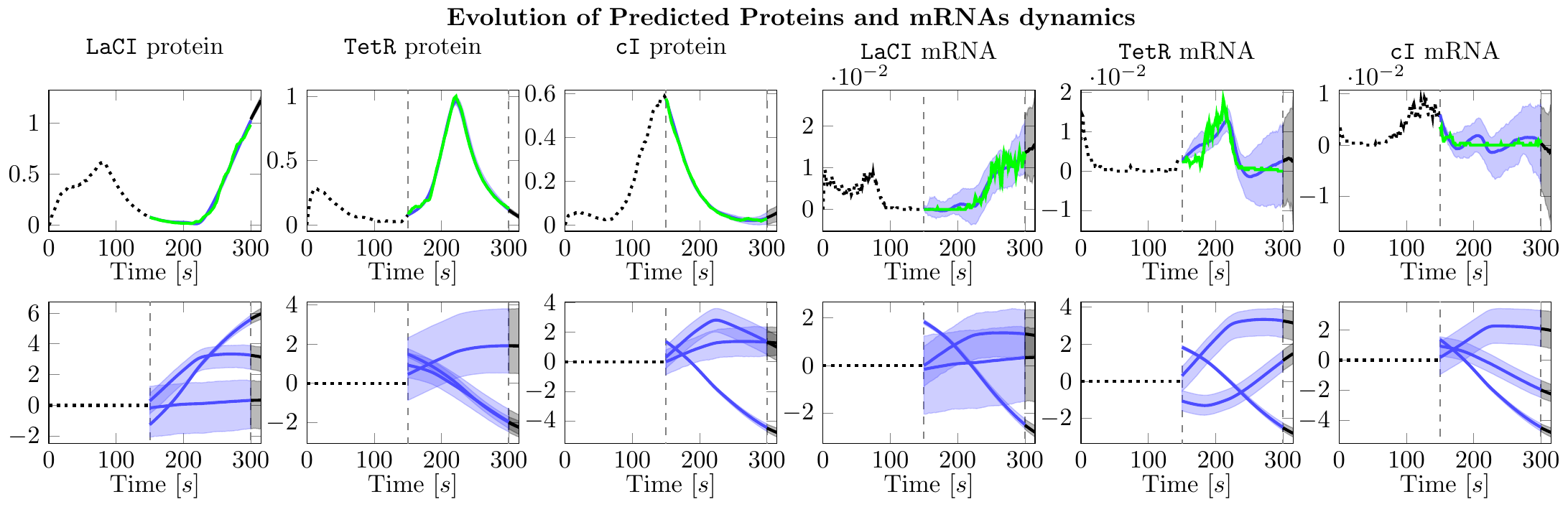}
    \vspace{-7mm}
    \caption{\footnotesize [Above] Reconstruction and extrapolation of protein and mRNA dynamics produced by Latent Neural SGDEs. Min--max intervals produced from multiple samples of the SDE decoder are indicated as shaded areas. [Below] Time--evolution of latent node features corresponding to reaction nodes connected to each of the protein and mRNA species in the reprissilator graph. The node reaction latent values evolve over time, modulating the reaction between protein species.}
    \label{fig:latentbois}
\end{figure*}

\paragraph{Results}
Figure~\ref{fig:latentbois} provides a visual inspection of protein and mRNA trajectory predictions produced by Latent Neural GDEs during testing. The model is able to reconstruct species concentration evolution, while calibrating and balancing decoder diffusion and drift to match the different characteristics of protein and mRNA dynamics. In black, we further highlight model extrapolations beyond the $300$ seconds regime, which shows an increase in model uncertainty. The attention coefficients of edges linking reaction nodes with the respective protein and mRNA species is shown in Figure~\ref{fig:attentionbois}. As the dynamics are unrolled by the decoder, the attention weights of the GAT \cite{velivckovic2017graph} layer present in the drift $f_{\cG}$ evolve over time, modulating reactions between species. Each protein and mRNA node is shown to correspond to six edge attention weights: three incoming and three outgoing.

\section{Related work}
Since the first seminal paper on Neural ODEs \citep{chen2018neural}, several attempts have been made at developing continuous variants of specific GNN models. Here, we provide a detailed comparison of Neural GDEs with other continuous formulations, highlighting in the process the need for a general framework systematically blending graphs and differential equations. \cite{sanchez2019hamiltonian} proposes using Graph Networks (GNs) \citep{battaglia2018relational} and ODEs to track Hamiltonian functions, obtaining \textit{Hamiltonian Graph Networks} (HGNs). The resulting model is evaluated on prediction tasks for conservative systems; an HGN can be expressed as a special case of a Neural GDE, and their application domain is fairly limited, being constrained to overly--restrictive Hamiltonian vector fields. In example, we evaluate Neural GDEs on multi--particle non--conservative systems subject to viscous friction, which are theoretically incompatible with HGNs. \citep{deng2019continuous} introduces a GNN version of continuous normalizing flows \citep{chen2018neural,grathwohl2018ffjord}, extending \citep{liu2019graph}, and deriving a continuous message passing scheme. However, the model is limited to the specific application of generative modeling, and no attempts at generalizing the formulation are made. Finally, \cite{xhonneux2019continuous} propose \textit{Continuous Graph Neural Networks} (CGNNs), a linear ODE formulation for message passing on graphs. CGNNs rely on rather strong assumptions, such as linearity or depth invariance, and are evaluated exclusively on datasets that can notably be tackled with linear GNNs \citep{wu2019simplifying}. Fashioning linear flows offers a closed--form solution of the model's output, though this is achieved at the cost of expressivity and generality. Indeed, CGNNs can also be regarded as a simple form of Neural GDEs. 

In contrast, our goal is to develop a system--theoretic framework for continuous--depth GNNs, validated through extensive experiments. To the best of our knowledge, no existing work offers a continuous--depth solution for dynamic graph prediction including latent and stochastic dynamics.
\section{Conclusion}
In this work we introduce \textit{neural graph differential equations} (Neural GDE), the continuous--depth counterpart to \textit{graph neural networks} (GNN) where the inputs are propagated through a continuum of GNN layers. Neural GDEs are designed to offer a data--driven modeling approach for \textit{dynamical networks}, whose dynamics are defined by a blend of discrete topological structures and differential equations. In sequential forecasting problems, Neural GDEs can accommodate irregular timestamps and track underlying continuous dynamics. Neural GDEs, including latent and stochastic variants, have been evaluated across applications, including traffic forecasting and prediction in genetic regulatory networks.

\clearpage
\bibliographystyle{abbrvnat}
\bibliography{main.bib}

\begin{thebibliography}{67}
\providecommand{\natexlab}[1]{#1}
\providecommand{\url}[1]{\texttt{#1}}
\expandafter\ifx\csname urlstyle\endcsname\relax
  \providecommand{\doi}[1]{doi: #1}\else
  \providecommand{\doi}{doi: \begingroup \urlstyle{rm}\Url}\fi

\bibitem[Andreasson et~al.(2014)Andreasson, Dimarogonas, Sandberg, and
  Johansson]{andreasson2014distributed}
M.~Andreasson, D.~V. Dimarogonas, H.~Sandberg, and K.~H. Johansson.
\newblock Distributed control of networked dynamical systems: Static feedback,
  integral action and consensus.
\newblock \emph{IEEE Transactions on Automatic Control}, 59\penalty0
  (7):\penalty0 1750--1764, 2014.

\bibitem[Atwood and Towsley(2016)]{atwood2016diffusion}
J.~Atwood and D.~Towsley.
\newblock Diffusion-convolutional neural networks.
\newblock In \emph{Advances in Neural Information Processing Systems}, pages
  1993--2001, 2016.

\bibitem[Battaglia et~al.(2018)Battaglia, Hamrick, Bapst, Sanchez-Gonzalez,
  Zambaldi, Malinowski, Tacchetti, Raposo, Santoro, Faulkner,
  et~al.]{battaglia2018relational}
P.~W. Battaglia, J.~B. Hamrick, V.~Bapst, A.~Sanchez-Gonzalez, V.~Zambaldi,
  M.~Malinowski, A.~Tacchetti, D.~Raposo, A.~Santoro, R.~Faulkner, et~al.
\newblock Relational inductive biases, deep learning, and graph networks.
\newblock \emph{arXiv preprint arXiv:1806.01261}, 2018.

\bibitem[Bruna et~al.(2013)Bruna, Zaremba, Szlam, and LeCun]{bruna2013spectral}
J.~Bruna, W.~Zaremba, A.~Szlam, and Y.~LeCun.
\newblock Spectral networks and locally connected networks on graphs.
\newblock \emph{arXiv preprint arXiv:1312.6203}, 2013.

\bibitem[Chen et~al.(2020)Chen, Lin, Li, Li, Zhou, and Sun]{chen2020measuring}
D.~Chen, Y.~Lin, W.~Li, P.~Li, J.~Zhou, and X.~Sun.
\newblock Measuring and relieving the over-smoothing problem for graph neural
  networks from the topological view.
\newblock In \emph{Proceedings of the AAAI Conference on Artificial
  Intelligence}, volume~34, pages 3438--3445, 2020.

\bibitem[Chen et~al.(2018)Chen, Rubanova, Bettencourt, and
  Duvenaud]{chen2018neural}
T.~Q. Chen, Y.~Rubanova, J.~Bettencourt, and D.~K. Duvenaud.
\newblock Neural ordinary differential equations.
\newblock In \emph{Advances in neural information processing systems}, pages
  6571--6583, 2018.

\bibitem[Cho et~al.(2014)Cho, Van~Merri{\"e}nboer, Gulcehre, Bahdanau,
  Bougares, Schwenk, and Bengio]{cho2014learning}
K.~Cho, B.~Van~Merri{\"e}nboer, C.~Gulcehre, D.~Bahdanau, F.~Bougares,
  H.~Schwenk, and Y.~Bengio.
\newblock Learning phrase representations using rnn encoder-decoder for
  statistical machine translation.
\newblock \emph{arXiv preprint arXiv:1406.1078}, 2014.

\bibitem[Defferrard et~al.(2016)Defferrard, Bresson, and
  Vandergheynst]{defferrard2016convolutional}
M.~Defferrard, X.~Bresson, and P.~Vandergheynst.
\newblock Convolutional neural networks on graphs with fast localized spectral
  filtering.
\newblock In \emph{Advances in neural information processing systems}, pages
  3844--3852, 2016.

\bibitem[Deng et~al.(2019)Deng, Nawhal, Meng, and Mori]{deng2019continuous}
Z.~Deng, M.~Nawhal, L.~Meng, and G.~Mori.
\newblock Continuous graph flow, 2019.

\bibitem[Dormand and Prince(1980)]{dormand1980family}
J.~R. Dormand and P.~J. Prince.
\newblock A family of embedded runge-kutta formulae.
\newblock \emph{Journal of computational and applied mathematics}, 6\penalty0
  (1):\penalty0 19--26, 1980.

\bibitem[Du et~al.(2020)Du, Futoma, and Doshi-Velez]{du2020model}
J.~Du, J.~Futoma, and F.~Doshi-Velez.
\newblock Model-based reinforcement learning for semi-markov decision processes
  with neural odes.
\newblock \emph{arXiv preprint arXiv:2006.16210}, 2020.

\bibitem[Dupont et~al.(2019)Dupont, Doucet, and Teh]{dupont2019augmented}
E.~Dupont, A.~Doucet, and Y.~W. Teh.
\newblock Augmented neural odes.
\newblock \emph{arXiv preprint arXiv:1904.01681}, 2019.

\bibitem[Elowitz and Leibler(2000)]{Elowitz2000}
M.~B. Elowitz and S.~Leibler.
\newblock A synthetic oscillatory network of transcriptional regulators.
\newblock \emph{Nature}, 403\penalty0 (6767):\penalty0 335--338, Jan 2000.
\newblock ISSN 1476-4687.
\newblock \doi{10.1038/35002125}.
\newblock URL \url{https://doi.org/10.1038/35002125}.

\bibitem[Gallicchio and Micheli(2019)]{gallicchio2019fast}
C.~Gallicchio and A.~Micheli.
\newblock Fast and deep graph neural networks, 2019.

\bibitem[Gasse et~al.(2019)Gasse, Ch{\'e}telat, Ferroni, Charlin, and
  Lodi]{gasse2019exact}
M.~Gasse, D.~Ch{\'e}telat, N.~Ferroni, L.~Charlin, and A.~Lodi.
\newblock Exact combinatorial optimization with graph convolutional neural
  networks.
\newblock \emph{arXiv preprint arXiv:1906.01629}, 2019.

\bibitem[Gillespie(2007)]{gillespie2007}
D.~T. Gillespie.
\newblock Stochastic simulation of chemical kinetics.
\newblock \emph{Annual Review of Physical Chemistry}, 58\penalty0 (1):\penalty0
  35--55, 2007.
\newblock \doi{10.1146/annurev.physchem.58.032806.104637}.
\newblock PMID: 17037977.

\bibitem[{Goebel} et~al.(2009){Goebel}, {Sanfelice}, and {Teel}]{goebel2008}
R.~{Goebel}, R.~G. {Sanfelice}, and A.~R. {Teel}.
\newblock Hybrid dynamical systems.
\newblock \emph{IEEE Control Systems Magazine}, 29\penalty0 (2):\penalty0
  28--93, 2009.

\bibitem[Grathwohl et~al.(2018)Grathwohl, Chen, Bettencourt, Sutskever, and
  Duvenaud]{grathwohl2018ffjord}
W.~Grathwohl, R.~T. Chen, J.~Bettencourt, I.~Sutskever, and D.~Duvenaud.
\newblock Ffjord: Free-form continuous dynamics for scalable reversible
  generative models.
\newblock \emph{arXiv preprint arXiv:1810.01367}, 2018.

\bibitem[Greydanus et~al.(2019)Greydanus, Dzamba, and
  Yosinski]{greydanus2019hamiltonian}
S.~Greydanus, M.~Dzamba, and J.~Yosinski.
\newblock Hamiltonian neural networks.
\newblock \emph{arXiv preprint arXiv:1906.01563}, 2019.

\bibitem[Jia and Benson(2019)]{jia2019neural}
J.~Jia and A.~R. Benson.
\newblock Neural jump stochastic differential equations.
\newblock \emph{arXiv preprint arXiv:1905.10403}, 2019.

\bibitem[Kidger et~al.(2020)Kidger, Morrill, Foster, and
  Lyons]{kidger2020neural}
P.~Kidger, J.~Morrill, J.~Foster, and T.~Lyons.
\newblock Neural controlled differential equations for irregular time series.
\newblock \emph{arXiv preprint arXiv:2005.08926}, 2020.

\bibitem[Kingma and Ba(2014)]{kingma2014adam}
D.~P. Kingma and J.~Ba.
\newblock Adam: A method for stochastic optimization.
\newblock \emph{arXiv preprint arXiv:1412.6980}, 2014.

\bibitem[Kipf and Welling(2016)]{kipf2016semi}
T.~N. Kipf and M.~Welling.
\newblock Semi-supervised classification with graph convolutional networks.
\newblock \emph{arXiv preprint arXiv:1609.02907}, 2016.

\bibitem[Kloeden et~al.(2012)Kloeden, Platen, and Schurz]{kloeden2012numerical}
P.~E. Kloeden, E.~Platen, and H.~Schurz.
\newblock \emph{Numerical solution of SDE through computer experiments}.
\newblock Springer Science \& Business Media, 2012.

\bibitem[Kunita(1997)]{kunita1997stochastic}
H.~Kunita.
\newblock \emph{Stochastic flows and stochastic differential equations},
  volume~24.
\newblock Cambridge university press, 1997.

\bibitem[Levie et~al.(2018)Levie, Monti, Bresson, and
  Bronstein]{levie2018cayleynets}
R.~Levie, F.~Monti, X.~Bresson, and M.~M. Bronstein.
\newblock Cayleynets: Graph convolutional neural networks with complex rational
  spectral filters.
\newblock \emph{IEEE Transactions on Signal Processing}, 67\penalty0
  (1):\penalty0 97--109, 2018.

\bibitem[Li et~al.(2020)Li, Wong, Chen, and Duvenaud]{li2020scalable}
X.~Li, T.-K.~L. Wong, R.~T. Chen, and D.~Duvenaud.
\newblock Scalable gradients for stochastic differential equations.
\newblock \emph{arXiv preprint arXiv:2001.01328}, 2020.

\bibitem[Li et~al.(2017)Li, Yu, Shahabi, and Liu]{li2017diffusion}
Y.~Li, R.~Yu, C.~Shahabi, and Y.~Liu.
\newblock Diffusion convolutional recurrent neural network: Data-driven traffic
  forecasting.
\newblock \emph{arXiv preprint arXiv:1707.01926}, 2017.

\bibitem[Li et~al.(2018{\natexlab{a}})Li, Vinyals, Dyer, Pascanu, and
  Battaglia]{li2018learning}
Y.~Li, O.~Vinyals, C.~Dyer, R.~Pascanu, and P.~Battaglia.
\newblock Learning deep generative models of graphs.
\newblock \emph{arXiv preprint arXiv:1803.03324}, 2018{\natexlab{a}}.

\bibitem[Li et~al.(2018{\natexlab{b}})Li, Chen, and
  Koltun]{li2018combinatorial}
Z.~Li, Q.~Chen, and V.~Koltun.
\newblock Combinatorial optimization with graph convolutional networks and
  guided tree search.
\newblock In \emph{Advances in Neural Information Processing Systems}, pages
  539--548, 2018{\natexlab{b}}.

\bibitem[Liu et~al.(2019)Liu, Kumar, Ba, Kiros, and Swersky]{liu2019graph}
J.~Liu, A.~Kumar, J.~Ba, J.~Kiros, and K.~Swersky.
\newblock Graph normalizing flows.
\newblock In \emph{Advances in Neural Information Processing Systems}, pages
  13556--13566, 2019.

\bibitem[Loshchilov and Hutter(2016)]{loshchilov2016sgdr}
I.~Loshchilov and F.~Hutter.
\newblock Sgdr: Stochastic gradient descent with warm restarts.
\newblock \emph{arXiv preprint arXiv:1608.03983}, 2016.

\bibitem[Lou et~al.(2020)Lou, Lim, Katsman, Huang, Jiang, Lim, and
  De~Sa]{lou2020neural}
A.~Lou, D.~Lim, I.~Katsman, L.~Huang, Q.~Jiang, S.-N. Lim, and C.~De~Sa.
\newblock Neural manifold ordinary differential equations.
\newblock \emph{arXiv preprint arXiv:2006.10254}, 2020.

\bibitem[Lu and Chen(2005)]{lu2005time}
J.~Lu and G.~Chen.
\newblock A time-varying complex dynamical network model and its controlled
  synchronization criteria.
\newblock \emph{IEEE Transactions on Automatic Control}, 50\penalty0
  (6):\penalty0 841--846, 2005.

\bibitem[Masid et~al.(2020)Masid, Ataman, and Hatzimanikatis]{Masid2020}
M.~Masid, M.~Ataman, and V.~Hatzimanikatis.
\newblock {Analysis of human metabolism by reducing the complexity of the
  genome-scale models using redHUMAN}.
\newblock \emph{Nature Communications}, 11\penalty0 (1):\penalty0 2821, 2020.
\newblock ISSN 2041-1723.
\newblock \doi{10.1038/s41467-020-16549-2}.
\newblock URL \url{https://doi.org/10.1038/s41467-020-16549-2}.

\bibitem[Massaroli et~al.(2020)Massaroli, Poli, Park, Yamashita, and
  Asama]{massaroli2020dissecting}
S.~Massaroli, M.~Poli, J.~Park, A.~Yamashita, and H.~Asama.
\newblock Dissecting neural odes.
\newblock \emph{arXiv preprint arXiv:2002.08071}, 2020.

\bibitem[Massaroli et~al.(2021)Massaroli, Poli, Peluchetti, Park, Yamashita,
  and Asama]{massaroli2021learning}
S.~Massaroli, M.~Poli, S.~Peluchetti, J.~Park, A.~Yamashita, and H.~Asama.
\newblock Learning stochastic optimal policies via gradient descent.
\newblock \emph{IEEE Control Systems Letters}, 2021.

\bibitem[Mathieu and Nickel(2020)]{mathieu2020riemannian}
E.~Mathieu and M.~Nickel.
\newblock Riemannian continuous normalizing flows.
\newblock \emph{arXiv preprint arXiv:2006.10605}, 2020.

\bibitem[Norcliffe et~al.(2020)Norcliffe, Bodnar, Day, Simidjievski, and
  Li{\`o}]{norcliffe2020second}
A.~Norcliffe, C.~Bodnar, B.~Day, N.~Simidjievski, and P.~Li{\`o}.
\newblock On second order behaviour in augmented neural odes.
\newblock \emph{arXiv preprint arXiv:2006.07220}, 2020.

\bibitem[{\O}ksendal(2003)]{oksendal2003stochastic}
B.~{\O}ksendal.
\newblock Stochastic differential equations.
\newblock In \emph{Stochastic differential equations}, pages 65--84. Springer,
  2003.

\bibitem[Oono and Suzuki(2019)]{oono2019graph}
K.~Oono and T.~Suzuki.
\newblock Graph neural networks exponentially lose expressive power for node
  classification, 2019.

\bibitem[Padgett and Ilie(2016)]{padgett2016tau}
J.~M.~A. Padgett and S.~Ilie.
\newblock An adaptive tau-leaping method for stochastic simulations of
  reaction-diffusion systems.
\newblock \emph{AIP Advances}, 6\penalty0 (3):\penalty0 035217, 2016.
\newblock \doi{10.1063/1.4944952}.
\newblock URL \url{https://doi.org/10.1063/1.4944952}.

\bibitem[Peluchetti and Favaro(2020)]{peluchetti2020infinitely}
S.~Peluchetti and S.~Favaro.
\newblock Infinitely deep neural networks as diffusion processes.
\newblock In \emph{International Conference on Artificial Intelligence and
  Statistics}, pages 1126--1136. PMLR, 2020.

\bibitem[Petneh{\'a}zi(2019)]{petnehazi2019recurrent}
G.~Petneh{\'a}zi.
\newblock Recurrent neural networks for time series forecasting.
\newblock \emph{arXiv preprint arXiv:1901.00069}, 2019.

\bibitem[Poli et~al.(2020)Poli, Massaroli, Yamashita, Asama, and
  Park]{poli2020torchdyn}
M.~Poli, S.~Massaroli, A.~Yamashita, H.~Asama, and J.~Park.
\newblock Torchdyn: A neural differential equations library, 2020.

\bibitem[Pontryagin et~al.(1962)Pontryagin, Mishchenko, Boltyanskii, and
  Gamkrelidze]{pontryagin1962mathematical}
L.~S. Pontryagin, E.~Mishchenko, V.~Boltyanskii, and R.~Gamkrelidze.
\newblock The mathematical theory of optimal processes.
\newblock 1962.

\bibitem[Rubanova et~al.(2019)Rubanova, Chen, and Duvenaud]{rubanova2019latent}
Y.~Rubanova, R.~T. Chen, and D.~Duvenaud.
\newblock Latent odes for irregularly-sampled time series.
\newblock \emph{arXiv preprint arXiv:1907.03907}, 2019.

\bibitem[Runge(1895)]{runge1895numerische}
C.~Runge.
\newblock {\"U}ber die numerische aufl{\"o}sung von differentialgleichungen.
\newblock \emph{Mathematische Annalen}, 46\penalty0 (2):\penalty0 167--178,
  1895.

\bibitem[Sanchez-Gonzalez et~al.(2018)Sanchez-Gonzalez, Heess, Springenberg,
  Merel, Riedmiller, Hadsell, and Battaglia]{sanchez2018graph}
A.~Sanchez-Gonzalez, N.~Heess, J.~T. Springenberg, J.~Merel, M.~Riedmiller,
  R.~Hadsell, and P.~Battaglia.
\newblock Graph networks as learnable physics engines for inference and
  control.
\newblock \emph{arXiv preprint arXiv:1806.01242}, 2018.

\bibitem[Sanchez-Gonzalez et~al.(2019)Sanchez-Gonzalez, Bapst, Cranmer, and
  Battaglia]{sanchez2019hamiltonian}
A.~Sanchez-Gonzalez, V.~Bapst, K.~Cranmer, and P.~Battaglia.
\newblock Hamiltonian graph networks with ode integrators.
\newblock \emph{arXiv preprint arXiv:1909.12790}, 2019.

\bibitem[Sandryhaila and Moura(2013)]{sandryhaila2013discrete}
A.~Sandryhaila and J.~M. Moura.
\newblock Discrete signal processing on graphs.
\newblock \emph{IEEE transactions on signal processing}, 61\penalty0
  (7):\penalty0 1644--1656, 2013.

\bibitem[Shuman et~al.(2013)Shuman, Narang, Frossard, Ortega, and
  Vandergheynst]{shuman2013emerging}
D.~I. Shuman, S.~K. Narang, P.~Frossard, A.~Ortega, and P.~Vandergheynst.
\newblock The emerging field of signal processing on graphs: Extending
  high-dimensional data analysis to networks and other irregular domains.
\newblock \emph{IEEE signal processing magazine}, 30\penalty0 (3):\penalty0
  83--98, 2013.

\bibitem[Smith and Topin(2019)]{smith2019super}
L.~N. Smith and N.~Topin.
\newblock Super-convergence: Very fast training of neural networks using large
  learning rates.
\newblock In \emph{Artificial Intelligence and Machine Learning for
  Multi-Domain Operations Applications}, volume 11006, page 1100612.
  International Society for Optics and Photonics, 2019.

\bibitem[Van Der~Schaft and Schumacher(2000)]{van2000introduction}
A.~J. Van Der~Schaft and J.~M. Schumacher.
\newblock \emph{An introduction to hybrid dynamical systems}, volume 251.
\newblock Springer London, 2000.

\bibitem[Vaswani et~al.(2017)Vaswani, Shazeer, Parmar, Uszkoreit, Jones, Gomez,
  Kaiser, and Polosukhin]{vaswani2017attention}
A.~Vaswani, N.~Shazeer, N.~Parmar, J.~Uszkoreit, L.~Jones, A.~N. Gomez,
  L.~Kaiser, and I.~Polosukhin.
\newblock Attention is all you need.
\newblock In \emph{Advances in neural information processing systems}, pages
  5998--6008, 2017.

\bibitem[Veli{\v{c}}kovi{\'c} et~al.(2017)Veli{\v{c}}kovi{\'c}, Cucurull,
  Casanova, Romero, Lio, and Bengio]{velivckovic2017graph}
P.~Veli{\v{c}}kovi{\'c}, G.~Cucurull, A.~Casanova, A.~Romero, P.~Lio, and
  Y.~Bengio.
\newblock Graph attention networks.
\newblock \emph{arXiv preprint arXiv:1710.10903}, 2017.

\bibitem[Weinan(2017)]{weinan2017proposal}
E.~Weinan.
\newblock A proposal on machine learning via dynamical systems.
\newblock \emph{Communications in Mathematics and Statistics}, 5\penalty0
  (1):\penalty0 1--11, 2017.

\bibitem[Wu et~al.(2019{\natexlab{a}})Wu, Zhang, Souza~Jr, Fifty, Yu, and
  Weinberger]{wu2019simplifying}
F.~Wu, T.~Zhang, A.~H.~d. Souza~Jr, C.~Fifty, T.~Yu, and K.~Q. Weinberger.
\newblock Simplifying graph convolutional networks.
\newblock \emph{arXiv preprint arXiv:1902.07153}, 2019{\natexlab{a}}.

\bibitem[Wu et~al.(2019{\natexlab{b}})Wu, Pan, Long, Jiang, and
  Zhang]{wu2019graph}
Z.~Wu, S.~Pan, G.~Long, J.~Jiang, and C.~Zhang.
\newblock Graph wavenet for deep spatial-temporal graph modeling.
\newblock \emph{arXiv preprint arXiv:1906.00121}, 2019{\natexlab{b}}.

\bibitem[Xhonneux et~al.(2019)Xhonneux, Qu, and Tang]{xhonneux2019continuous}
L.-P.~A. Xhonneux, M.~Qu, and J.~Tang.
\newblock Continuous graph neural networks.
\newblock \emph{arXiv preprint arXiv:1912.00967}, 2019.

\bibitem[Yan et~al.(2018)Yan, Xiong, and Lin]{yan2018spatial}
S.~Yan, Y.~Xiong, and D.~Lin.
\newblock Spatial temporal graph convolutional networks for skeleton-based
  action recognition.
\newblock In \emph{Thirty-Second AAAI Conference on Artificial Intelligence},
  2018.

\bibitem[Y{\i}ld{\i}z et~al.(2019)Y{\i}ld{\i}z, Heinonen, and
  L{\"a}hdesm{\"a}ki]{yildiz2019ode}
{\c{C}}.~Y{\i}ld{\i}z, M.~Heinonen, and H.~L{\"a}hdesm{\"a}ki.
\newblock Ode$^{2}$vae: Deep generative second order odes with bayesian neural
  networks.
\newblock \emph{arXiv preprint arXiv:1905.10994}, 2019.

\bibitem[You et~al.(2018)You, Ying, Ren, Hamilton, and
  Leskovec]{you2018graphrnn}
J.~You, R.~Ying, X.~Ren, W.~L. Hamilton, and J.~Leskovec.
\newblock Graphrnn: Generating realistic graphs with deep auto-regressive
  models.
\newblock \emph{arXiv preprint arXiv:1802.08773}, 2018.

\bibitem[You et~al.(2019)You, Wu, Barrett, Ramanujan, and
  Leskovec]{you2019g2sat}
J.~You, H.~Wu, C.~Barrett, R.~Ramanujan, and J.~Leskovec.
\newblock G2sat: Learning to generate sat formulas.
\newblock In \emph{Advances in neural information processing systems}, pages
  10553--10564, 2019.

\bibitem[Yu et~al.(2018)Yu, Yin, and Zhu]{yu2018spatio}
B.~Yu, H.~Yin, and Z.~Zhu.
\newblock Spatio-temporal graph convolutional networks: A deep learning
  framework for traffic forecasting.
\newblock In \emph{Proceedings of the 27th International Joint Conference on
  Artificial Intelligence (IJCAI)}, 2018.

\bibitem[Zhao et~al.(2018)Zhao, Chen, and Cho]{zhao2018deep}
X.~Zhao, F.~Chen, and J.-H. Cho.
\newblock Deep learning for predicting dynamic uncertain opinions in network
  data.
\newblock In \emph{2018 IEEE International Conference on Big Data (Big Data)},
  pages 1150--1155. IEEE, 2018.

\bibitem[Zhuang and Ma(2018)]{zhuang2018dual}
C.~Zhuang and Q.~Ma.
\newblock Dual graph convolutional networks for graph-based semi-supervised
  classification.
\newblock In \emph{Proceedings of the 2018 World Wide Web Conference}, pages
  499--508. International World Wide Web Conferences Steering Committee, 2018.

\end{thebibliography}
\clearpage
\newpage
\onecolumn
\rule[0pt]{\columnwidth}{3pt}
\begin{center}
\huge{\bf{Continuous--Depth Neural Models for Dynamic Graph Prediction} \\
\emph{Supplementary Material}}
\end{center}
\vspace*{3mm}
\rule[0pt]{\columnwidth}{1pt}
\vspace*{-.5in}
\vspace*{1in}

\appendix
\vspace*{-3cm}
\addcontentsline{toc}{section}{}
\part{}
\parttoc
%
\section{Neural Graph Differential Equations}
\paragraph{Notation}
Let $\Nat$ be the set of natural numbers and $\R$ the set of reals. Indices of arrays and matrices are reported as superscripts in round brackets.

Let $\V$ be a finite set with $|\V| = n$ whose element are called \textit{nodes} and let $\E$ be a finite set of tuples of $\V$ elements. Its elements are called \textit{edges} and are such that $\forall e_{ij}\in\E,~e_{ij} = (v_i,v_j)$ and $v_i,v_j\in\V$. A graph $\G$ is defined as the collection of nodes and edges, i.e. $\G := (\V,\E)$. The \textit{adjacency} matrix $A\in\R^{n\times n}$ of a graph is defined as
\[
    A^{(ij)} = \left\{
        \begin{matrix*}[l]
        1 & e_{ij}\in\E\\
        0 & e_{ij}\not\in\E
        \end{matrix*}
        \right.~.
\]
If $\G$ is an \textit{attributed graph}, the \textit{feature vector} of each $v\in\V$ is $x_v\in\R^{n_x}$. All the feature vectors are collected in a matrix $X\in\R^{n\times n_x}$. Note that often, the features of graphs exhibit temporal dependency, i.e. $X := X_t$.

\subsection{Standalone Neural GDE formulation}
For clarity and as an easily accessible reference, we include below a general formulation table for Neural GDEs
\begin{CatchyBox}{Neural GDEs}
    \begin{minipage}[h]{0.35\linewidth}
            \begin{equation*}\label{eq:1}
            {
                \left\{
                \begin{aligned}
                    \dot Z_t &= f_{\cG}\lc t, Z_t, \theta_t\rc \\
                    Z_0 &= \ell_{\cG}^x (X)\\
                    \hat{Y}_t &= \ell_{\cG}^y (Z_t) 
                \end{aligned}
                \right.~~
                t\in \mathcal{T}}
            \end{equation*}
            \hfill
    \end{minipage}
    \hfill
    \begin{minipage}[h]{.61\linewidth}\small
        \centering
        \begin{tabular}{r|c|l}
            Input node features & $X$&$\R^{n\times n_x}$\\\hline
            Output & $Y$&$\R^{n\times n_y}$\\\hline
            Graph&$\cG$ &$n$ nodes\\\hline
            Latent node features & $Z$&$\R^{n\times \nZ}$\\\hline
            Parameters & $\theta$&$\R^{\nT}$\\\hline
            Vector Field & $f_{\G}$ & $\R\times\R^{n\times \nZ}\times\R^{\nT}\rightarrow \R^{n\times\nZ}$\\\hline
            Input Network & $\ell_{\cG}^x$ & $\R^{n\times\nX}\rightarrow\R^{n\times\nZ}$\\\hline
            Output Network & $\ell_{\cG}^y$ & $\R^{n\times\nZ}\rightarrow\R^{n\times\nY}$
        \end{tabular}
    \end{minipage}
\end{CatchyBox}
Note that the system provided in \eqref{eq:hybrid} can serve as a similar reference for the spatio--temporal case.
\subsection{Computational Overhead}
As is the case for other models sharing the continuous--depth formulation \citep{chen2018neural}, the computational overhead required by Neural GDEs depends mainly by the numerical methods utilized to solve the differential equations. We can define two general cases for \textit{fixed--step} and \textit{adaptive--step} solvers.
\paragraph{Fixed--step}
In the case of fixed--step solvers of \textit{k--th} order e.g \textit{Runge--Kutta--k} \citep{runge1895numerische}, the time complexity is $O(nk)$ where $n := S/\epsilon$ defines the number of steps necessary to cover $[0, S]$ in fixed--steps of $\epsilon$.\hfill

\begin{wrapfigure}[15]{r}{0.52\textwidth}
    \centering
     \includegraphics[width=1\linewidth]{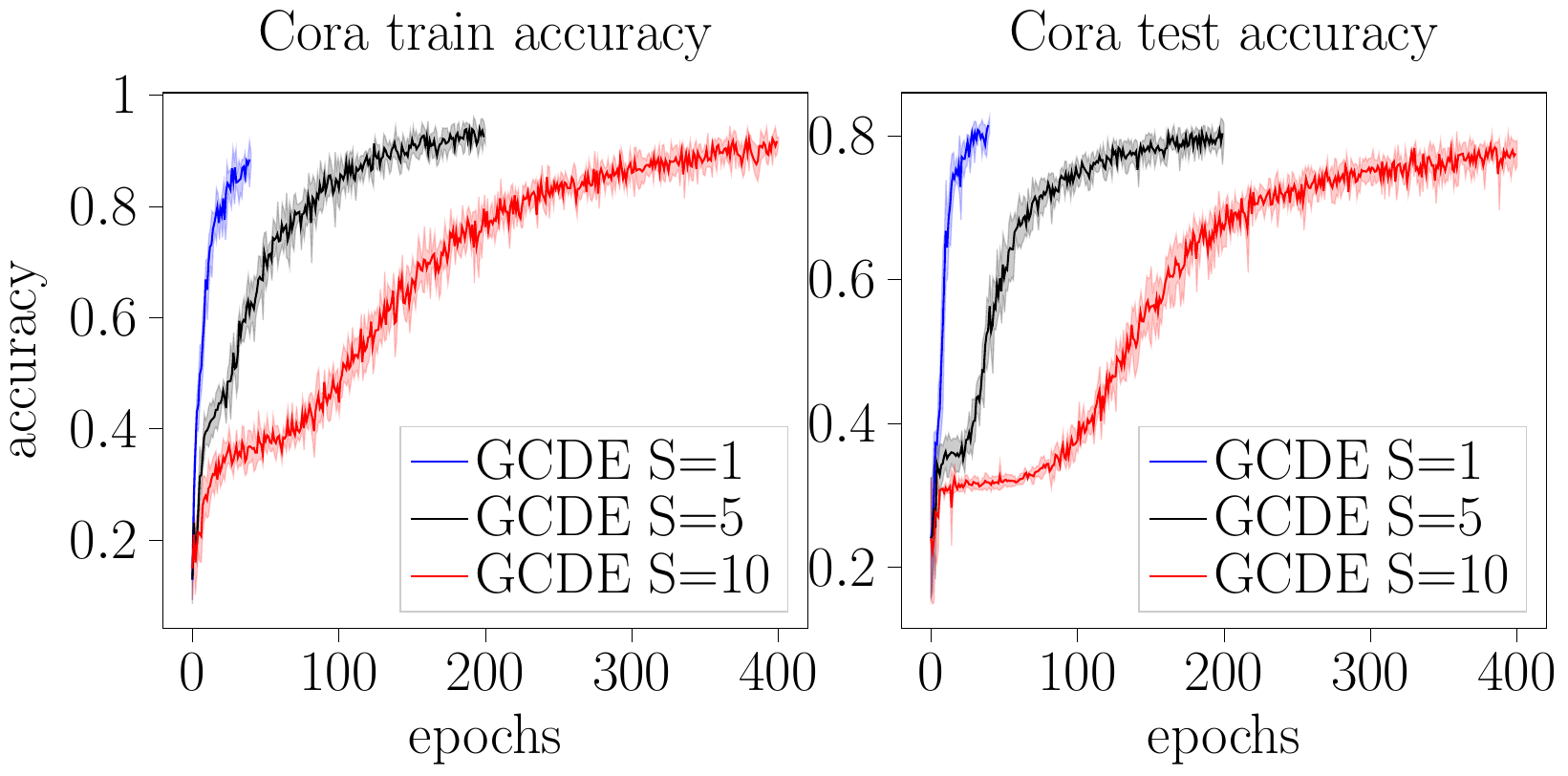}
    \caption{Cora accuracy of Neural GCDE models with different integration intervals $S$. Longer integration does not result in loss of performance, only an increase in number of training iterations required for convergence.}
    \label{fig:trk4}
\end{wrapfigure}
\paragraph{Adaptive--step}
For general adaptive--step solvers, computational overhead ultimately depends on the error tolerances. While worst--case computation is not bounded \citep{dormand1980family}, a maximum number of steps can usually be set algorithmically. 
\subsection{Oversmoothing in Neural GDEs}

For each upper limit of integration in $[1, 5, 10]$, we train $100$ \textit{Neural Graph Convolutional Differential Equations} (Neural GCDEs) models on standard static dataset Cora and report average metrics, along with $1$ standard deviation confidence intervals in Figure \ref{fig:trk4}. Neural GCDEs are shown to be resilient to these changes; however, with longer integration they require more training epochs to achieve comparable accuracy. This result suggests that Neural GDEs are immune to node oversmoothing \citep{oono2019graph}, as their differential equation is not, unless by design, stable, and thus does not reach an equilibrium state for node representations. Repeated iteration of discrete GNN layers, on the other hand, has been empirically observed to converge to fixed points which make downstream tasks less performant \citep{chen2020measuring}.
\subsection{Additional Neural GDEs}
\paragraph{Message passing Neural GDEs}  Let us consider a single node $v\in\V$ and define the set of neighbors of $v$ as $\N(v): = \{u\in\V~:~(v,u)\in\E \lor ~~(u,v)\in\E\}$. Message passing neural networks (MPNNs) perform a spatial--based convolution on the node $v$ as
\begin{equation}\label{eq:MPNN}
    z^{(v)}{(s+1)} = \phi\left[z^{(v)}(s), \sum_{u\in\N(v)}m\lc{z^{(v)}(s),z^{(u)}(s)}\rc\right],
\end{equation}
where, in general, $z^{v}(0) = x_v$ while $\phi$ and $m$ are functions with trainable parameters. For clarity of exposition, let $\phi(x,y) := x + g(y)$ where $g$ is the actual parametrized function. The previous system (\ref{eq:MPNN}) becomes
\begin{equation}
    z^{(v)}{(s+1)} = z^{(v)}(s) + g\left[\sum_{u\in\N(v)}m\lc{z^{(v)}(s),z^{(u)}(s)}\rc\right],
\end{equation}
and its continuous--depth counterpart, \textit{graph message passing differential equation} (GMDE) is:
\begin{equation*}
    \dot z^{(v)}{(s)} = f^{(v)}_{\tt MPNN}(Z, \theta) := g\left[\sum_{u\in\N(v)}m\lc{z^{(v)}(s),z^{(u)}}(s)\rc\right].
\end{equation*}

\paragraph{Attention Neural GDEs} Graph attention networks (GATs) \citep{velivckovic2017graph} perform convolution on the node $v$ as
\begin{equation}
    z^{(v)}{(s+1)} = \sigma\lc\sum_{u\in\N(v)\cup v}{\alpha_{vu}W(s)z^{(u)}}(s)\rc.
\end{equation}
Similarly, to GCNs, a \textit{virtual} skip connection can be introduced allowing us to define the \textit{neural graph attention differential equation} (Neural GADE):
\begin{equation*}
    \dot z^{(v)}{(s)} = f^{(v)}_{\tt GAT}(Z(s),\theta):= \sigma\lc\sum_{u\in\N(v)\cup v}{\alpha_{vu}Wz^{(u)}}(s)\rc, \theta:=\vect(W)
\end{equation*}
where $\alpha_{vu}$ are attention coefficient which can be computed following \citep{velivckovic2017graph}.

The attention operator is introduced within drift functions of Latent Neural GDE. Here, visualizing the time--evolution of attention coefficients allows an inspection of neighbouring nodes importance in driving particular node dynamics.
\subsection{Hybrid Adjoints}
Let us recall the back--propagated adjoint gradients for Hybrid Neural GDEs defined in Section \ref{sec:3.1}
\[
    \begin{aligned}
        \frac{\dd L_\theta}{d\theta} &= \frac{\partial L_\theta}{\partial \theta}-\int_{\cT} \left<\lambda_t, \frac{\partial f_{\cG_t}(t, \Phi_{t}(Z_0),\theta)}{\partial\theta}\right>\dd t\\
        \dot \lambda_t &\in F(t, Z_t, \lambda_t)~~~~~t \in \cT\\
        \lambda^+_t & \in G(X_t, Z_t, \lambda_t)~~~~~t\in \cT_e\\
        \lambda_{t_K} &= \frac{\partial c_\theta}{\partial Z_{t_K}} 
    \end{aligned}
\]
where
\[ 
    \begin{aligned}
        F(t, Z_t, \lambda_t):=\left\{F_k:t\in[t_k,t_{k+1}]\Rightarrow F_k=-\frac{\partial f_{\cG_{t_k}}}{\partial Z}\lambda_t\right\}\\
        G(X_t, Z_t, \lambda_t):=\left\{G_k:t=t_k\Rightarrow G_k= \lambda_t + \frac{\partial \ell^y_{\G_{t_k}}}{\partial Z_t}+ \frac{\partial c_\theta}{\partial Z_t}\right\}.
    \end{aligned}
\]
The above formulation can be derived using the \textit{hybrid inclusions} formalism \citep{goebel2008} by extending the results of \citep{jia2019neural}. In fact, in presence of discontinuities (jumps) in the state the during the forward integration of the (hybrid) differential equation, the adjoint state dynamics becomes itself an hybrid dynamical system with jumps
\[
    \lambda_t^+ =\lambda_t + \frac{\partial \ell^y_{\G_{t_k}}}{\partial Z_t}+ \frac{\partial c_\theta}{\partial Z_t}~\text{ if }~ t=t_k\in\cT_e.
\]
Due to the time--varying nature of the graph $\G_t$ and, consequently, of the \textit{flow} and \textit{jump} maps $f_{\G_{t_k}}$, $\ell_{\G_{t_k}}^y$ in the Hybrid Neural GDE, then also the flow and jump maps of the adjoint systems will be different in each interval $[t_k, t_{k+1}]$. The hybrid inclusion representation formalizes this discrete--continuous time--varying nature of the dynamics.
\section{Spatio--Temporal Neural GDEs}
We include a complete description of GCGRUs to clarify the model used in our experiments.

\subsection{GCGRU Cell}
Following GCGRUs \citep{zhao2018deep}, we perform an instantaneous jump of $Z$ at each time $t_{k}$ using the next input features $X_{t_{k}}$. Let $L_{\G_{t_k}}$ be the graph Laplacian of graph $\G_{t_k}$, which can computed in several ways \citep{bruna2013spectral,defferrard2016convolutional,levie2018cayleynets, zhuang2018dual}. Then, let
\begin{equation}
\begin{aligned} 
H&:=\sigma\left(L_{\G_{t_k}}X_{t_k}W_{xz} + L_{\G_{t_k}}Z W_{hz}\right), \\
R &:=\sigma\left(L_{\G_{t_k}}X_{t_k}W_{xr} + L_{\G_{t_k}}Z W_{hr}\right), \\ 
\tilde{Z} &:=\tanh \left(L_{\G_{t_k}}X_{t_k}W_{xh} + L_{\G_{t_k}}\left(R \odot Z\right)W_{hh}\right). 
\end{aligned}
\end{equation}
Finally, the \textit{post--jump} node features are obtained as
\begin{equation}
\begin{aligned} 
Z^+ &={\tt GCGRU}(Z,X_t):=H \odot Z+(\mathbb{1}-H) \odot \tilde~{Z} \end{aligned}
\end{equation}
where $W_{xz},~W_{hz},~W_{xr},~W_{hr},~W_{xh},~W_{hh}$ are matrices of trainable parameters and $\sigma$ is the standard sigmoid activation and $\mathbb{1}$ is all--ones matrix of suitable dimensions.
\section{Additional experimental details}

\paragraph{Computational resources}
We carried out all experiments on a cluster of 2x24GB NVIDIA\textsuperscript{\textregistered} {\tt RTX} $3090$ and CUDA 11.2. The models were trained on GPU. All experiments can be run on a single GPU, as memory requirements never exceeded memory capacity ($\leq ~10${\tt GB}).

\begin{figure*}[t]
    \centering
    \includegraphics[width=1\linewidth]{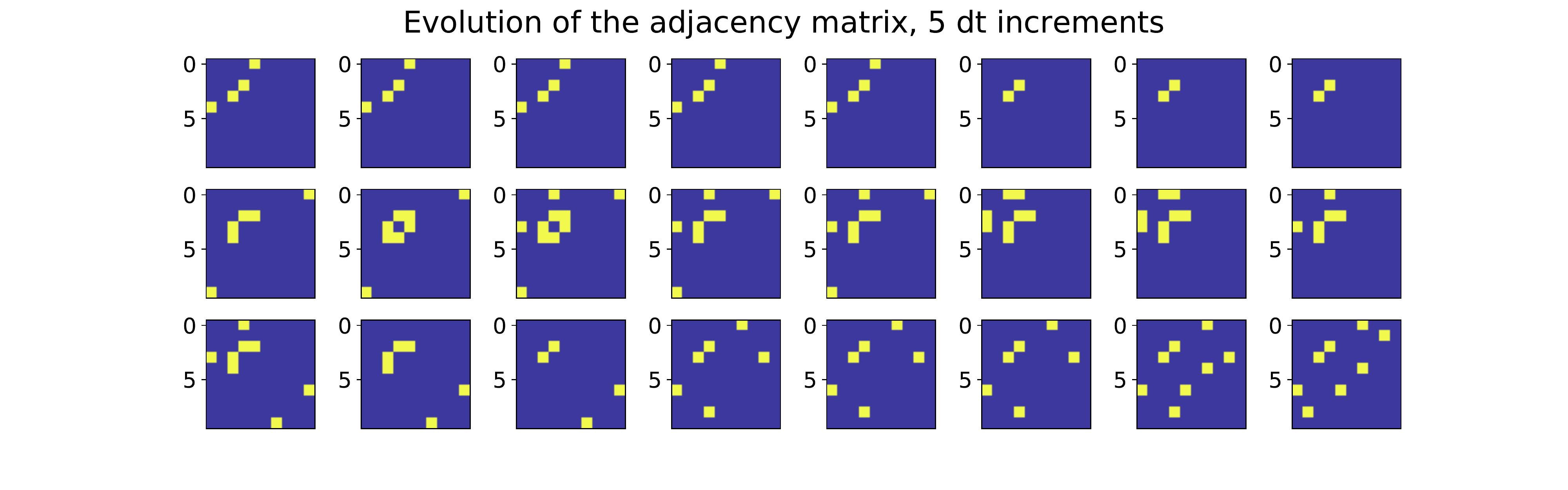}
    \vspace{-5mm}
    \caption{Snapshots of the evolution of adjacency matrix $A_t$ throughout the dynamics of the multi--particle system. Yellow indicates the presence of an edge and therefore a reciprocal force acting on the two bodies}
    \label{fig:adj}
\end{figure*}
\subsection{Multi--Agent System Dynamics}
\begin{figure}[t]
    \centering
     \includegraphics[width=0.5\linewidth]{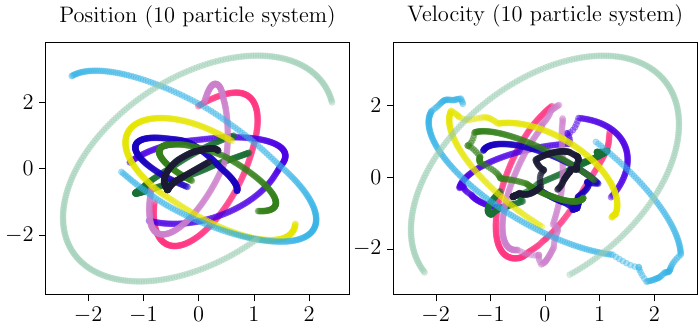}
    \caption{Example position and velocity trajectories of the multi--particle system.}
    \label{fig:ma_traj}
\end{figure}

\paragraph{Dataset}
Let us consider a planar multi agent system with states $x_i$ ($i = 1,\dots,n$) and second--order dynamics:
\begin{align*}
    \ddot x_i &= -x_i - \sum_{j\in\N_i}f_{ij}(x_i,x_j,\dot x_i,\dot x_j),
\end{align*}
where 
\begin{align*}
    f_{ij} = -\left[\alpha\lc \|x_i - x_j\| - r \rc + \beta\frac{\langle\dot x_i - \dot x_j, x_i - x_j\rangle}{\|x_i - x_j\|}\right]{n}_{ij},\\
    {n}_{ij} = \frac{x_i - x_j}{\|x_i - x_j\|},\quad \alpha,~\beta,~r > 0,
\end{align*}
and
\[
    \quad \N_i := \left\{j:2\|x_i-x_j\|\leq r \land j\not = i\right\}.
\]
The force $f_{ij}$ resembles the one of a spatial spring with drag interconnecting the two agents. The term $-x_i$, is used instead to stabilize the trajectory and avoid the "explosion" of the phase--space. Note that $f_{ij} = -f_{ji}$. The adjaciency matrix $A_t$ is computed along a trajectory

\[
    A_t^{(ij)} = 
    \left\{\begin{matrix*}[l]
        1 & 2\|x_i(t)-x_j(t)\|\leq r\\
        0 & \text{otherwise}
    \end{matrix*}\right.,
\]
which indeed results to be symmetric, $A_t = A_t^\top$ and thus yields an undirected graph. Figure \ref{fig:adj} visualizes an example trajectory of $A_t$, and Figure \ref{fig:ma_traj} trajectories of the system.
\begin{table*}
\centering
\setlength\tabcolsep{3pt}
\begin{tabular}[t]{lrrrrrrr}  
\toprule
Model & MAPE$_{1}$ & MAPE$_{3}$ & MAPE$_{5}$ & MAPE$_{10}$ & MAPE$_{15}$ & MAPE$_{20}$ & MAPE$_{50}$ \\
\midrule
Static & $26.12$ & $160.56$ & $197.20$ & $ 235.21$ & $261.56$ & $275.60$ & $360.39$ \\
Neural ODE  & $ 26.12$ & $52.26$ & $92.31$ & $ 156.26$ & $238.14$ & $301.85$ & $668.47$  \\
Neural GDE & $13.53$ & $15.22$ & $18.76$ & $27.76$ & $33.90$ & $42.22$ & $77.64$ \\
Neural GDE--II & $13.46$ & $14.75$ & $17.81$ & $27.77$ & $32.28$ & $40.64$ & $73.75$  \\
\bottomrule
\end{tabular}
\caption{Mean MAPE results across the 10 multi--particle dynamical system experiments. MAPE$_i$ indicates results for $i$ extrapolation steps on the full test trajectory.}
\label{tab:mape_table}
\end{table*}
We collect a single rollout with $T = 5$, $dt = 1.95\cdot10^{-3}$ and $n = 10$. The particle radius is set to $r = 1$.

\leavevmode
\paragraph{Architectural details}

\begin{wrapfigure}[25]{r}{0.52\textwidth}
    \centering
    \includegraphics[scale=0.4]{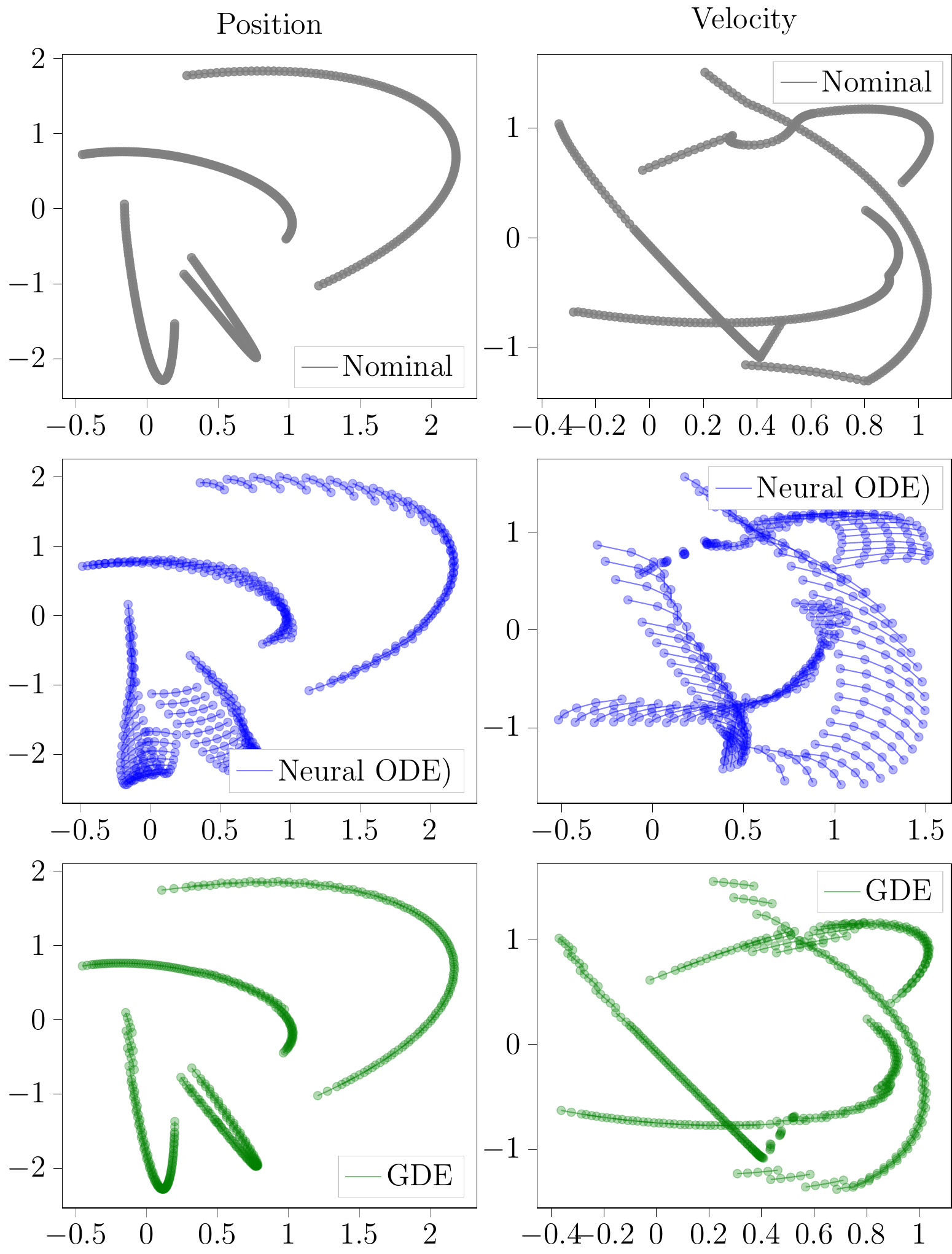}
    \caption{Test extrapolation, $5$ steps. Trajectory predictions of Neural ODEs and Neural GDEs. Extrapolation is terminated after 5 steps and the nominal state is fed to the model.}
    \label{fig:extrapo_traj}
\end{wrapfigure}
Node feature vectors are $4$ dimensional, corresponding to the dimension of the state, i.e. position and velocity. Neural ODEs and \textit{Static} share an architecture made up of 3 fully--connected layers: $4n$, $8n$, $8n$, $4n$ where $n = 10$ is the number of nodes. The last layer is linear. We evaluated different hidden layer dimensions: $8n$, $16n$, $32n$ and found $8n$ to be the most effective. Similarly, the architecture of first order Neural GCDEs is composed of 3 GCN layers: $4$, $16$, $16$, $4$. Second--order Neural GCDEs, on the other hand, are augmented by $4$ dimensions: $8$, $32$, $32$, $8$. We experimented with different ways of encoding the adjacency matrix $A$ information into Neural ODEs and $Static$ but found that in all cases it lead to worse performance. 
\paragraph{Additional results}
We report in Figure~\ref{fig:extrapo_traj} test extrapolation predictions of $5$ steps for Neural GDEs and the various baselines. Neural ODEs fail to track the system, particularly in regions of the state space where interaction forces strongly affect the dynamics. Neural GDEs, on the other hand, closely track both positions and velocities of the particles.

\subsection{Traffic Forecasting}
\begin{figure*}[h!]
    \centering
    \begin{tikzpicture}

\definecolor{color0}{rgb}{0.12156862745098,0.466666666666667,0.705882352941177}

\begin{groupplot}[group style={group size=3 by 1}]
\nextgroupplot[
width = 4.5 cm,
height = 3cm,
tick align=outside,
tick pos=left,
title={Keep probability 30\%},
x grid style={white!69.0196078431373!black},
xlabel={Delta Time Stamp},
xmin=0, xmax=22,
xtick style={color=black},
y grid style={white!69.0196078431373!black},
ylabel={Frequency},
ymin=0, ymax=0.916538592896175,
ytick style={color=black}
]
\draw[draw=none,fill=color0] (axis cs:1,0) rectangle (axis cs:3,0.253164556962025);
\draw[draw=none,fill=color0] (axis cs:3,0) rectangle (axis cs:5,0.124472573839662);
\draw[draw=none,fill=color0] (axis cs:5,0) rectangle (axis cs:7,0.0638185654008439);
\draw[draw=none,fill=color0] (axis cs:7,0) rectangle (axis cs:9,0.0313818565400844);
\draw[draw=none,fill=color0] (axis cs:9,0) rectangle (axis cs:11,0.01292194092827);
\draw[draw=none,fill=color0] (axis cs:11,0) rectangle (axis cs:13,0.00659282700421941);
\draw[draw=none,fill=color0] (axis cs:13,0) rectangle (axis cs:15,0.00395569620253165);
\draw[draw=none,fill=color0] (axis cs:15,0) rectangle (axis cs:17,0.00210970464135021);
\draw[draw=none,fill=color0] (axis cs:17,0) rectangle (axis cs:19,0.000791139240506329);
\draw[draw=none,fill=color0] (axis cs:19,0) rectangle (axis cs:21,0.000791139240506329);

\nextgroupplot[
width = 4.5cm,
height = 3cm,
tick align=outside,
tick pos=left,
title={Keep probability 50\%},
x grid style={white!69.0196078431373!black},
xlabel={Delta Time Stamp},
xmin=0.45, xmax=12.55,
xtick style={color=black},
y grid style={white!69.0196078431373!black},
ymin=0, ymax=0.916538592896175,
ytick style={color=black}
]
\draw[draw=none,fill=color0] (axis cs:1,0) rectangle (axis cs:2.1,0.682529743268629);
\draw[draw=none,fill=color0] (axis cs:2.1,0) rectangle (axis cs:3.2,0.117265327033643);
\draw[draw=none,fill=color0] (axis cs:3.2,0) rectangle (axis cs:4.3,0.0537940456537826);
\draw[draw=none,fill=color0] (axis cs:4.3,0) rectangle (axis cs:5.4,0.0281778334376957);
\draw[draw=none,fill=color0] (axis cs:5.4,0) rectangle (axis cs:6.5,0.0128081061080435);
\draw[draw=none,fill=color0] (axis cs:6.5,0) rectangle (axis cs:7.6,0.00626174076393237);
\draw[draw=none,fill=color0] (axis cs:7.6,0) rectangle (axis cs:8.7,0.00313087038196618);
\draw[draw=none,fill=color0] (axis cs:8.7,0) rectangle (axis cs:9.8,0.00313087038196619);
\draw[draw=none,fill=color0] (axis cs:9.8,0) rectangle (axis cs:10.9,0.00113849832071498);
\draw[draw=none,fill=color0] (axis cs:10.9,0) rectangle (axis cs:12,0.000853873740536233);

\nextgroupplot[
width = 4.5cm,
height = 3cm,
tick align=outside,
tick pos=left,
title={Keep probability 70\%},
x grid style={white!69.0196078431373!black},
xlabel={Delta Time Stamp},
xmin=0.6, xmax=9.4,
xtick style={color=black},
y grid style={white!69.0196078431373!black},
ymin=0, ymax=0.916538592896175,
ytick style={color=black}
]
\draw[draw=none,fill=color0] (axis cs:1,0) rectangle (axis cs:1.8,0.872893897996357);
\draw[draw=none,fill=color0] (axis cs:1.8,0) rectangle (axis cs:2.6,0.255009107468124);
\draw[draw=none,fill=color0] (axis cs:2.6,0) rectangle (axis cs:3.4,0.0836748633879781);
\draw[draw=none,fill=color0] (axis cs:3.4,0) rectangle (axis cs:4.2,0.0281762295081967);
\draw[draw=none,fill=color0] (axis cs:4.2,0) rectangle (axis cs:5,0);
\draw[draw=none,fill=color0] (axis cs:5,0) rectangle (axis cs:5.8,0.00711520947176684);
\draw[draw=none,fill=color0] (axis cs:5.8,0) rectangle (axis cs:6.6,0.00199225865209472);
\draw[draw=none,fill=color0] (axis cs:6.6,0) rectangle (axis cs:7.4,0.000569216757741348);
\draw[draw=none,fill=color0] (axis cs:7.4,0) rectangle (axis cs:8.2,0.000284608378870674);
\draw[draw=none,fill=color0] (axis cs:8.2,0) rectangle (axis cs:9,0.000284608378870674);
\end{groupplot}

\end{tikzpicture}
    \caption{Distribution of deltas between timestamps $t_{k+1} - t_k$ in the undersampled dataset. The time scale of required predictions varies greatly during the task.}
    \label{fig:traffic_delta_ts}
\end{figure*}
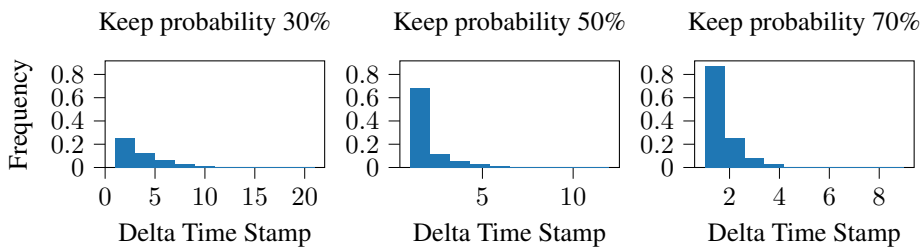

\paragraph{Dataset and metrics}
The timestamp differences between consecutive graphs in the sequence varies due to undersampling. The distribution of timestamp deltas (5 minute units) for the three different experiment setups (30\%, 50\%, 70\% undersampling) is shown in Figure \ref{fig:traffic_delta_ts}.

As a result, GRU takes 230 dimensional vector inputs (228 sensor observations + 2 additional features) at each sequence step. Both GCGRU and GCDE--GRU graph inputs with and 3 dimensional node features (observation + 2 additional feature). The additional time features are excluded for the loss computations.
We include MAPE and RMSE test measurements, defined as follows:
\begin{equation}
    \text{MAPE}(\yb, \hat{\yb}) = \frac{100 \%}{pT}\norm{\sum_{t=1}^{T}(y_t - \hat{y}_t) \oslash y_t}_1,
\end{equation}
where $\yb, \text{and} \ \hat{\yb} \in \mathbb R^p $ is the set of vectorized target and prediction of models respectively. $\oslash$ and $\norm{\cdot}_1$ denotes Hadamard division and the 1-norm of vector. 

\begin{align*}
    \text{RMSE}(\yb, \hat{\yb}) &= \frac{1}{p}\norm{\sqrt{\frac{1}{T}\sum_{t=1}^{T}  (y_t - \hat{y}_t)^2}}_1,
\end{align*}

where $(\cdot)^2$ and $\sqrt{\cdot}$ denotes the element-wise square and square root of the input vector, respectively. $\yb_t \ \text{and} \ \hat{\yb}_t$ denote the target and prediction vector. 
\paragraph{Architectural details}
We employed two baseline models for contextualizing the importance of key components of GCDE--GRU. \textit{GRUs} architectures are equipped with 1 GRU layer with hidden dimension 50 and a 2 layer fully--connected head to map latents to predictions. \textit{GCGRUs} employ a GCGRU layer with 46 hidden dimension and a 2 layer fully--connected head. Lastly, GCDE--GRU shares the same architecture GCGRU with the addition of the flow $f_{\G}$ tasked with evolving the hidden features between arrival times. $f_{\G}$ is parametrized by 2 GCN layers, one with tanh activation and the second without activation. ReLU is used as the general activation function.

\paragraph{Training hyperparameters}
All models are trained for 40 epochs using Adam~\citep{kingma2014adam} with $lr = 10^{-2}$. We schedule $lr$ by using cosine annealing method \citep{loshchilov2016sgdr} with $T_0=10$. The optimization is carried out by minimizing the \textit{mean square error} (MSE) loss between predictions and corresponding targets. 

\begin{figure}[t]
    \centering
    \includegraphics{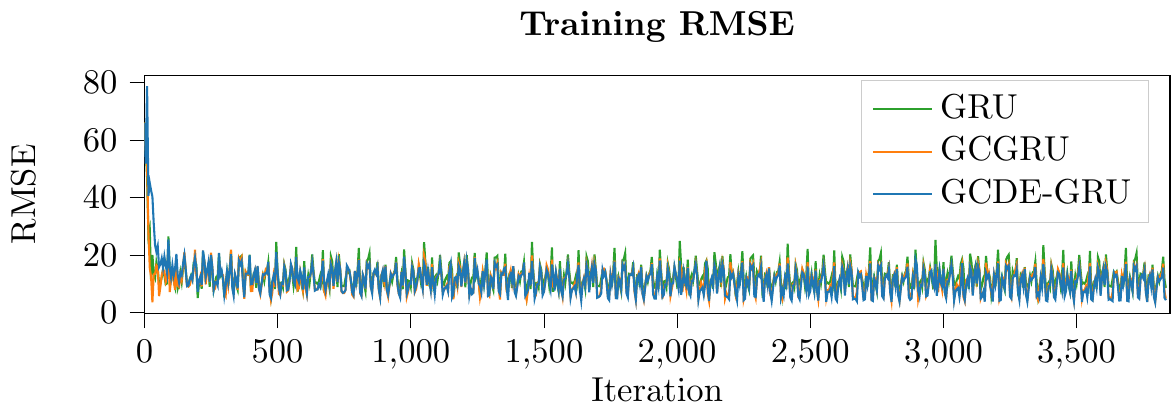}
    \caption{Traffic data \textit{unsmoothed} RMSE during a 50\% undersampling training experiment.}
    \label{fig:traffic_train}
\end{figure}

\paragraph{Additional results}
Training curves of the models are presented in the Fig \ref{fig:traffic_train}. All of models achieved nearly 13 in RMSE during training and fit the dataset. However, due to the lack of dedicated spatial modeling modules, GRUs were unable to generalize to the test set and resulted in a mean value prediction. 
\begin{figure}[t]
    \centering
    \includegraphics[width=0.5\linewidth]{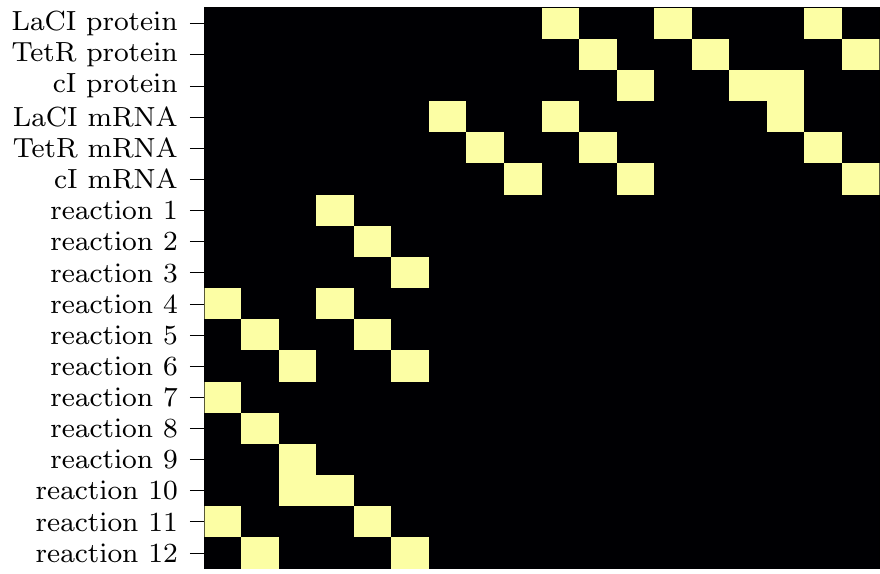}
    \caption{Adjacency matrix of Latent Neural GDE decoders on the repressilator prediction task.}
    \label{fig:adj_repri}
\end{figure}

\subsection{Repressilator Reconstruction}
\paragraph{Dataset}
We construct a dataset of $10$ training and $10$ test trajectories of a \citep{Elowitz2000} Elowitz--Leibler reprissilator system. The data is collected by running stochastic simulations using the $\tau$-leaping method \citep{gillespie2007,padgett2016tau} from an initial condition of $[0, 0, 0, 0, 20, 0]$, where the first three components are concentrations of {\tt LaCI}, {\tt TetR} and {\tt cI} proteins respectively, whereas the last three are concentrations of mRNAs. In the main text, we qualitatively inspect reconstruction capability of the Latent Neural GDE across test trajectories. The samples conditioned on the first $150$ seconds of each simulation closely follow the solutions obtained via $\tau$--leaping. The adjacency matrix is constructed as shown in Fig.~\ref{fig:adj_repri}, where the three species of proteins and are disconnected, able to interact only through reaction and mRNA nodes.

\paragraph{Architectural details}
We employ a Latent Neural GDE model for the reprissilator prediction task. The decoder is comprised of a \textit{neural stochastic graph differential equation} (Neural GSDE) with drift and diagonal diffusion functions constructed as detailed in Table \ref{tab:gde_arch}. The encoder is constructed with a two layers of \textit{temporal convolutions} (TCNs) and ReLU activations, applied to the $6$ protein and mRNA trajectories.   
\begin{table}
\centering
\setlength\tabcolsep{3pt}
\begin{tabular}[t]{lrrr}  
\toprule
Layer & Input dim. & Output dim. & Activation \\
\midrule
GCN--1 & $1$ & $3$ & Tanh \\
GAT & $3$ & $3$ & Tanh  \\
GCN--2 & $3$ & $1$ & None \\
\bottomrule

\end{tabular}
\vspace{2mm}
\caption{General architecture for Neural GSDEs drift $f_{\cG}$ and diffusion $g_{\cG}$ functions in decoders of Latent Neural GDEs. The drift function $g_{\cG}$ is equipped with an additional sigmoid activation at the end to enforce non--negativity.}
\label{tab:gde_arch}
\end{table}

\paragraph{Training hyperparameters}
Latent Neural GDEs are trained for $1000$ epochs using Adam \citep{kingma2014adam}. We use a one cycle scheduling policy 
\citep{smith2019super} for the learning rate with maximum at $10^{-2}$ at $300$ epochs and a minimum of $4\cdot 10^{-4}$. The Neural GSDE decoder is solved using an adaptive Euler--Heun \citep{kloeden2012numerical} scheme with tolerances set to $10^{-3}$. All data is min--max normalized before being fed into the model.

\end{document}